\documentclass[pdflatex, sn-basic,iicol]{sn-jnl}

\jyear{2021}%

\theoremstyle{thmstyleone}%
\setcitestyle{authoryear,open={(},close={)}}

\theoremstyle{thmstyletwo}%

\theoremstyle{thmstylethree}%

\usepackage{caption}
\usepackage{flushend}

\bibpunct[, ]{(}{)}{;}{a}{,}{,}
\begin{document}

\title[SHARP]{SHARP: Shape-Aware Reconstruction of People in Loose Clothing}


\author{\fnm{Sai Sagar} \sur{Jinka}}\email{jinka.sagar@research.iiit.ac.in}

\author{\fnm{Astitva Srivastava}}\email{astitva.srivastava@research.iiit.ac.in}

\author{\fnm{Chandradeep} \sur{Pokhariya}}\email{chandradeep.pokhariya@research.iiit.ac.in}

\author{\fnm{Avinash} \sur{Sharma }}\email{asharma@iiit.ac.in}
\author{\fnm{P.J.} \sur{Narayanan}}\email{pjn@iiit.ac.in}
\affil*[1]{\orgdiv{Centre for Visual Information Technology}, \orgname{IIIT Hyderabad}, \orgaddress{\street{Gachibowli}, \city{Hyderabad}, \postcode{500032}, \state{Telangana}, \country{India}}}


\abstract{
Recent advancements in deep learning have enabled 3D human body reconstruction from a monocular image, which has broad applications in multiple domains. In this paper, we propose SHARP (\textbf{SH}ape \textbf{A}ware \textbf{R}econstruction of \textbf{P}eople in loose clothing), a novel end-to-end trainable network that accurately recovers the 3D geometry and appearance of humans in loose clothing from a monocular image. SHARP uses a sparse and efficient fusion strategy to combine parametric body prior with a non-parametric 2D representation of clothed humans. The parametric body prior enforces geometrical consistency on the body shape and pose, while the non-parametric representation models loose clothing and handles self-occlusions as well. We also leverage the sparseness of the non-parametric representation for faster training of our network while using losses on 2D maps. Another key contribution is \emph{3DHumans}, our new life-like dataset of 3D human body scans with rich geometrical and textural details. We evaluate SHARP on 3DHumans and other publicly available datasets, and show superior qualitative and quantitative performance than existing state-of-the-art methods.}

\keywords{3D human body reconstruction, parametric and non-parametric methods, monocular image, deep learning}

\maketitle

\section{Introduction}\label{sec:introduction}

Image-based 3D reconstruction of humans in loose clothing is an interesting and challenging open problem in computer vision. It has several applications in the domains of fashion, AR/VR, sports and healthcare. Traditional stereo/multi-view (including RGB and depth sensor) based reconstruction solutions
 \citep{gall2009motion,shotton2011real,wei2012accurate,baak2013data,newcombe2015dynamicfusion,dou2016fusion4d,bogo2017dynamic} typically require studio environments with controlled lighting and multiple synchronized and calibrated cameras. 
\begin{figure*}[t!]
             \centering
            \includegraphics[width=\linewidth]{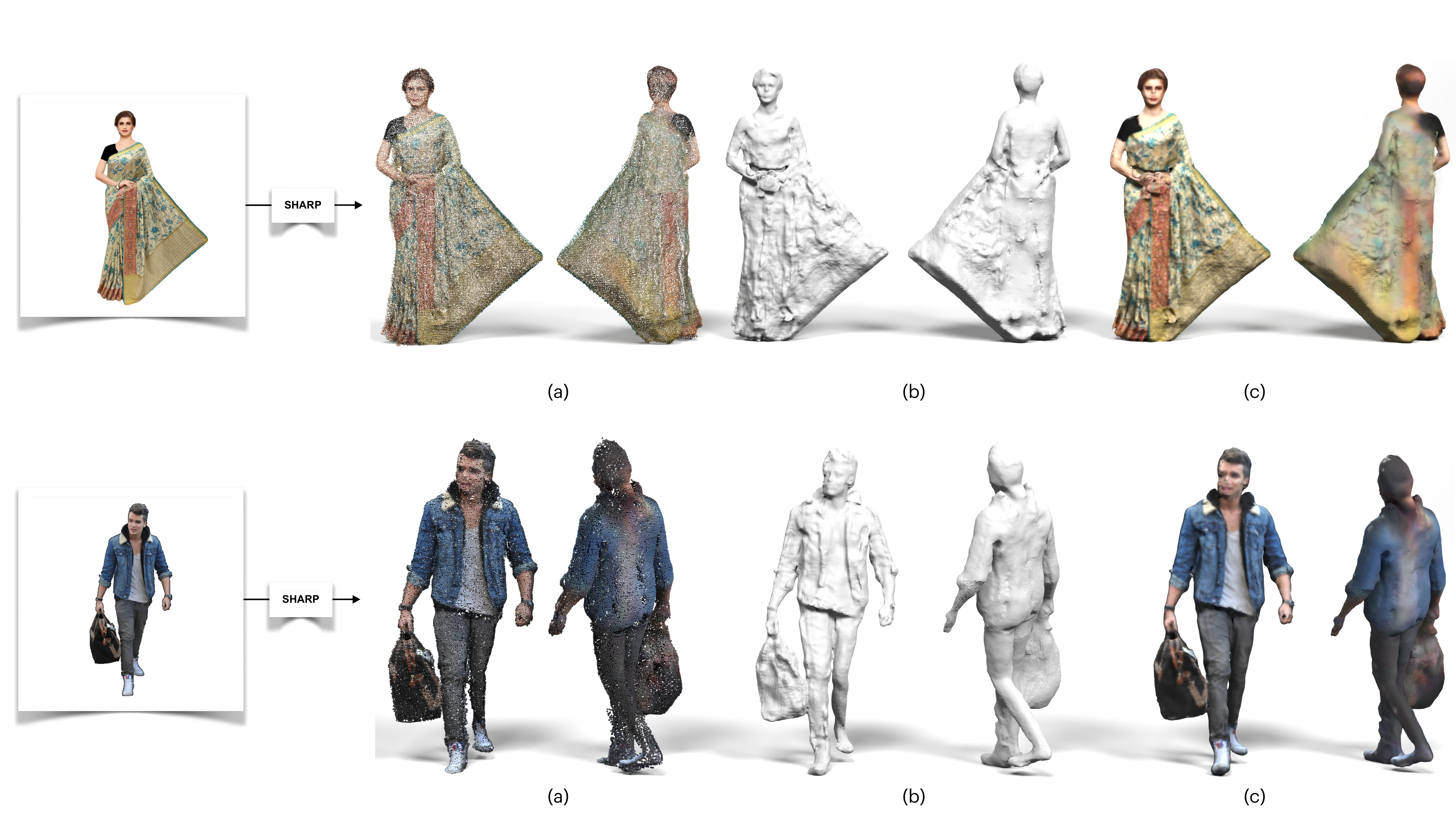}
            \caption{Results of our method on in-the-wild images. Point cloud, uncolored and colored mesh is shown in (a), (b) \& (c), respectively.}
            \label{fig:teaser}
        \end{figure*}
Thus, recent approaches have shifted their focus on in-the-wild 3D reconstruction of humans. 

With the advent of deep learning models, significant interest has garnered around 3D reconstruction from a monocular image \citep{kanazawa2018end,varol2018bodynet,habermann2020deepcap}, which is an ill-posed problem. Challenges like self-occlusions, arbitrary viewpoints and clothing occlusions make the scenario more complicated.
One class of existing deep learning solutions attempts to fit a parametric body model like SMPL \citep{SMPL:2015} to a monocular input image by learning from image features  \citep{kanazawa2018end,guler2018densepose,omran2018neural,lin2021mesh,kolotouros2021probabilistic}. SMPL prediction is improved when Multi-view input images are provided as shown in \citep{liang2019shape}. However, such parametric SMPL mesh does not capture geometrical details owing to person-specific appearance and clothing.
The other class of non-parametric reconstruction techniques pose no such body prior constraints  \citep{saito2019pifu,saito2020pifuhd,natsume2019siclope,varol2018bodynet,bhatnagar2020loopreg,venkat2018deep} and hence can potentially handle loose clothing scenarios. 

In particular, the recent implicit function learning models, like PIFu \citep{saito2019pifu} and PIFuHD \citep{saito2020pifuhd}, estimate voxel occupancy by utilizing pixel-aligned RGB image features computed by projecting 3D points onto the input image. However, the pixel-aligned features suffer from depth ambiguity as multiple 3D points are projected to the same pixel. Another interesting work, Geo-PIFu  \citep{he2020geo} attempted to refine implicit function estimation by combining volumetric features and pixel-aligned features together to resolve local feature ambiguity. As an alternate representation for 3D objects/scenes, some of the recent works model scenes as multiple (depth) plane images (MPIs) \citep{single_view_mpi} in camera frustum. 3D human body reconstruction has also been attempted in the same vein by predicting front and back depth maps in \citep{mouldinghumans,smith2019facsimile}. However, the front-back representation fails to handle self-occlusions caused by body parts. 

In our recent work peeledhuman \citep{jinka2020peeledhuman}, we introduced \emph{PeeledHuman}; a novel non-parametric shape representation of the human body to address the self-occlusion problem. 
PeeledHuman representation encodes the 3D human body shape as a set of depth and RGB peel maps. Depth (and RGB) peeling is performed by ray-tracing on the 3D body mesh and extending each ray beyond its first intersection to obtain the peel maps. 
This provides an elegant, sparse 2D encoding of body shape, which inherently addresses the self-occlusion problem. However, the non-parametric approaches do not explicitly seek to impose global body shape consistency and hence, produces implausible body shape and pose.

The aforementioned problems can be addressed by introducing a body shape prior while reconstructing humans in loose clothing. The volume-to-volume translation network proposed in DeepHuman  \citep{zheng2019deephuman} attempts to combine image features with the SMPL prior in a volumetric representation. ARCH \citep{huang2020arch,he2021arch++} proposed to induce a human body prior by sampling points around a template SMPL mesh before evaluating occupancy labels for each point.
However, sampling around the canonical SMPL surface is insufficient to reconstruct humans with articulated poses in loose clothing. Similarly, PaMIR \citep{zheng2021pamir} proposes to voxelize SMPL body and feed it as an input to the network, which conditions the implicit function around the SMPL feature volume. 
However, volumetric feature estimation is still computationally expensive and is limited by the resolution.
Moreover, in PaMIR,  texture and geometry cannot be inferred in an end-to-end fashion and require two separate networks.
Additionally, all these existing SMPL prior-based methods do not effectively exploit the rich surface representation as they either voxelize or sample points around the SMPL surface.

The continuous surface representation provided by SMPL prior is valuable as it models the natural curvature of body parts which cannot be easily recovered with non-parametric methods. Some of the existing methods have been successfully shown to deform SMPL surfaces locally to accommodate relatively tight clothing scenarios  \citep{alldieck19cvpr,bhatnagar2019mgn,patel20tailornet,alldieck2019tex2shape,lahner2018deepwrinkles,zhu2019detailed}. 
Nevertheless, they fail to handle loose clothing scenarios, as the surface of garments can also have complex geometrical structures that are only partially dependent on the underlying body shape and pose, where non-parametric methods have mainly been successful.
Interestingly, we can retain the best of these two approaches by deforming SMPL surface locally
while reconstructing the remaining surface details (loose clothing) with no body prior constraints. 
More specifically, one can decouple the reconstruction of 3D clothed body surface into two complementary partial reconstruction tasks: (a) to recover the person-specific body surface details by locally deforming the SMPL prior, (b) to recover the remaining surface details of the loose clothing that cannot be recovered by just deforming the SMPL prior.

In regard to the representation of the 3D surface, while implementing the above two tasks, PeeledHuman representation seems to be a good choice owing to its sparse encoding of a 3D surface into 2D maps. More importantly, such representation also enables a seamless fusion of the two partial reconstructions due to the spatially aligned nature of these maps. 3D geometry can be extracted from PeeledHuman representation by simply back-projecting the peel maps to generate point cloud. Recent works \citep{Ma:CVPR:2021, POP:ICCV:2021} have shown that point clouds are a good way to model clothing deformations arising from articulated pose.

Thus, this work proposes SHARP, a novel 3D body reconstruction method that can successfully handle significantly loose clothing, self-occlusions and arbitrary viewpoints.
SHARP takes SMPL body encoded in PeeledHuman representation \citep{jinka2020peeledhuman}, aligned to the input image as a prior to the reconstruction framework. The \emph{SMPL prior peel maps}, along with the monocular RGB image, is fed as the input to our framework, which initially predicts \emph{residual peel maps}, \emph{auxiliary peel maps}, along with \emph{RGB peel maps}.
Here, the residual peel maps represent the pixel-wise depth offsets from SMPL prior peel maps in the view direction. On the other hand, auxiliary peel maps model the complementary geometrical details of the surface, which are not handled by residual peel maps.
Subsequently, predicted residual and auxiliary peel maps are fused to obtain \emph{fused peel maps}, capturing the geometry of the unified clothed body.
The final fused peel maps, along with predicted RGB peel maps are back-projected to obtain the colored point cloud. We finally recover the mesh after minimal post-processing of the corresponding point cloud followed by meshification using Poisson Surface Reconstruction\citep{kazhdan2006poisson}.
The fused peel maps can model arbitrarily loose clothing and can handle accessories (e.g., bags) as well, as shown in \autoref{fig:teaser}. 
Unlike other existing methods that use adversarial loss and 3D Chamfer loss, the proposed problem formulation enables our network to learn only with $L_1$ losses on 2D maps, which reduces the training time. Since, the clothed human body can be recovered in the form of point cloud directly from the back-projection of final fused peel maps, the inference time is also significantly reduced. We have described layerwise back-projection in greater detail in \textbf{section 2} of the supplementary draft.

Additionally, many state-of-the-art methods for reconstructing 3D human bodies \citep{saito2019pifu,saito2020pifuhd,zheng2021pamir,natsume2019siclope,huang2020arch} train their models on expensive commercial datasets which are not publicly available. These datasets have 3D human body scans which resemble real humans. This data helps the learning-based models to generalize well on unseen real-world scenarios.
Unfortunately, the majority of existing datasets available in the public domain \citep{bhatnagar2019mgn, zheng2019deephuman,bertiche2020Cloth3D,tiwari2020sizer} either consist of 3D body models in relatively tighter clothing, lack high-frequency geometrical \& texture details, or are synthetic in nature.
Recently, THUman2.0 \citep{yu2021function4d} dataset released in the public domain has high-quality 3D body scans captured using a dense DSLR rig. Although they provide human scans with relatively loose clothing styles,  their data lacks significantly loose garment types which occlude the lower body completely, e.g., long-skirt/tunic/saree. Moreover, the dataset is reconstructed with the multi-camera setup which has its known limitations.
To bridge these gaps, we collected \emph{\textbf{3DHumans}}, a dataset of 3D human body scans with a wide variety of clothing styles and varied poses using a commercial structured-light sensor (accurate up to 0.5mm). We are able to retain high-frequency geometrical and textural details, as shown in \autoref{fig:our_dataset}. We also benchmark some of the SOTA methods on this dataset and report superior performance of our method. 
To summarize, our contributions are:
\begin{enumerate}
\item{We propose SHARP, a novel approach to fuse parametric and non-parametric shape representations for reconstructing 3D body model in loose clothing from an input monocular (RGB) image.}
\item{ Our proposed end-to-end learnable encoder-decoder framework infers color and geometrical details of body shape in a single forward pass at a lower inference time as compare to SOTA methods.}
\item{We collected \emph{\textbf{3DHumans}}, a dataset of 3D human body scans that has a wide variety of clothing and body poses with rich textural and geometrical details. The dataset will be released in the public domain to further accelerate the research.}
\end{enumerate}

\section{Related Work}\label{sec:related_work}

\textbf{Parametric Body Fitting.} 
 Estimating the 3D parametric human body models, like SMPL \citep{SMPL:2015}, SMPL-X \citep{pavlakos2019expressive}, SCAPE \citep{anguelov2005scape} etc.,  from a monocular image using deep learning methods \citep{Bogo:ECCV:2016, kanazawa2018end} has achieved a great success with robust performance. In particular, HMR \citep{kanazawa2018end} proposes to regress SMPL parameters while minimizing re-projection loss with the known 2D joints.  
Different priors have been used to refine the parametric estimates as in \citep{varol2017learning,omran2018neural,kolotouros2019learning,kanazawa2019learning,kolotouros2021probabilistic,lin2021mesh}.
Despite these approaches being computationally efficient, they lack realistic human appearance and clothing details. Methods for modelling details like hair/cloth/skin by estimating offsets from SMPL vertex have been proposed, but they work on very tight clothing and can not model the loose clothing deformation arising from pose . \citep{bhatnagar2019mgn,venkat2019humanmeshnet,kolotouros2019convolutional}. \\  
\\
\textbf{Non-parametric Body Reconstruction:} Recovering 3D human body from multi-camera setup requires traditional techniques like voxel carving, triangulation, multi-view stereo, shape-from-X  \citep{azevedo20093d,dou2016fusion4d,bogo2017dynamic,mulayim2003silhouette}. Stereo cameras and consumer RGBD sensors are highly susceptible to noise.
In the domain of deep learning, initially, voxel methods gained popularity as 3D voxels are a natural extension to 2D pixels \citep{venkat2018deep,varol2018bodynet,zheng2019deephuman}. 
SiCloPe \citep{natsume2019siclope} estimates human body silhouettes in novel views to recover underlying 3D shape from 2D contours. Recently, implicit function learning methods for human body reconstruction became popular, which use pixel-aligned features to learn neural implicit function over a discrete occupancy grid~\citep{saito2019pifu,saito2020pifuhd}. However, these methods suffer from sampling redundancy as they have to sample points in a grid to infer the surface, majority of which do not lie on the actual surface. They also suffer from depth ambiguity as multiple 3D points map to the same pixel-aligned feature.
Animating clothed humans with template garments is proposed in \citep{corona2021smplicit}. However, this method cannot produce the textural details of the garments.
In our recent work peeledhuman~\citep{jinka2020peeledhuman}, we proposed a sparse 2D representation of 3D surface by estimating and storing the intersection of the surface with ray.\citep{mildenhall2020nerf} where it samples points along the camera ray to evaluate RGB$ \sigma$ on these samples.\\
\\
\textbf{Prior-based Non-Parametric Body Reconstruction:} ARCH \citep{huang2020arch,he2021arch++} learns a deep implicit function by sampling points around the 3D clothed body in the canonical space. But, the transformation of the clothed mesh from canonical space to arbitrary space is done by learning SMPL-based skinning weights which can not handle the deformation of the loose clothing. These methods rely on large scale dataset of 3D human scans to train the model, and suffer from reconstruction errors and weak generalization capability although demonstrating good results.
\citep{bhatnagar2020combining} also proposes to combine strengths of parametric and non-parametric models. However, it takes input as sparse point cloud which is difficult to obtain in-the-wild settings.
 Geo-PIFu \citep{he2020geo} utilizes structure-aware latent voxel features, along with pixel-aligned features to learn a neural implicit function. PaMIR \citep{zheng2021pamir} learns a deep implicit function conditioned on the features which are a combination of 2D features obtained from image and 3D features obtained from the SMPL body volume. However, voxel features are computationally expensive and of low resolution. DeepHuman \citep{zheng2019deephuman} leverages dense semantic representations from SMPL as an additional input. Nevertheless, similar to Geo-PIFu, DeepHuman is also a volumetric-regression based approach and hence, incurs a high computational cost. Moreover, similar to PIFu, these deep implicit methods require separate networks for learning geometry and texture.\\
\\
\textbf{3D Human Body Datasets:} Deep-Learning based 3D human body reconstruction solutions rely on the data available at hand. Not only the shear amount of samples, but the quality of geometry and texture is also important in order to drive the learning.
Many 3D human body datasets have been proposed, some of which only contain body-shape information, while some also include clothing details on top of it. TOSCA \citep{bronstein2008numerical} dataset contains synthetic meshes of fixed topology with artist-defined deformations. SHREC \citep{li2012shrec} and FAUST \citep{bogo2014faust} provide meshes and deformation models created by an artist that cannot reproduce what we find in the real world. 
BUFF \citep{zhang2017detailed} contains 3D scans with relatively richer geometry details, but the number of subjects, poses and clothing style is very limited and not sufficient to generalize deep learning models. Another synthetic dataset CLOTH3D \citep{bertiche2020Cloth3D} incorporates loose clothing by draping 3D modeled garments on SMPL in Blender. It has a wide variety of clothing styles, but due to the nature of SMPL body model, details like hair and skin are absent.  
THUman1.0 \citep{zheng2019deephuman} dataset provides a large number of human meshes with varied poses and subjects. However, the texture quality is low and cannot mimic real-world subjects. 
SIZER \citep{tiwari2020sizer} dataset provides real scans of 100 subjects, wearing garments in 4 different sizes of 10 fixed garments classes. But all the scans are in A-pose which is insufficient for a deep learning model to generalize to different poses.
THUman2.0 \citep{yu2021function4d} dataset provides a large number of high-quality textured meshes of different subjects in various poses. It also incorporates varied clothing styles and high-frequency geometrical details like hair and wrinkles etc. However, loose wrapped clothing styles, which completely occlude the full body, are still absent.
\begin{figure*}
            \centering
            \includegraphics[width=\linewidth]{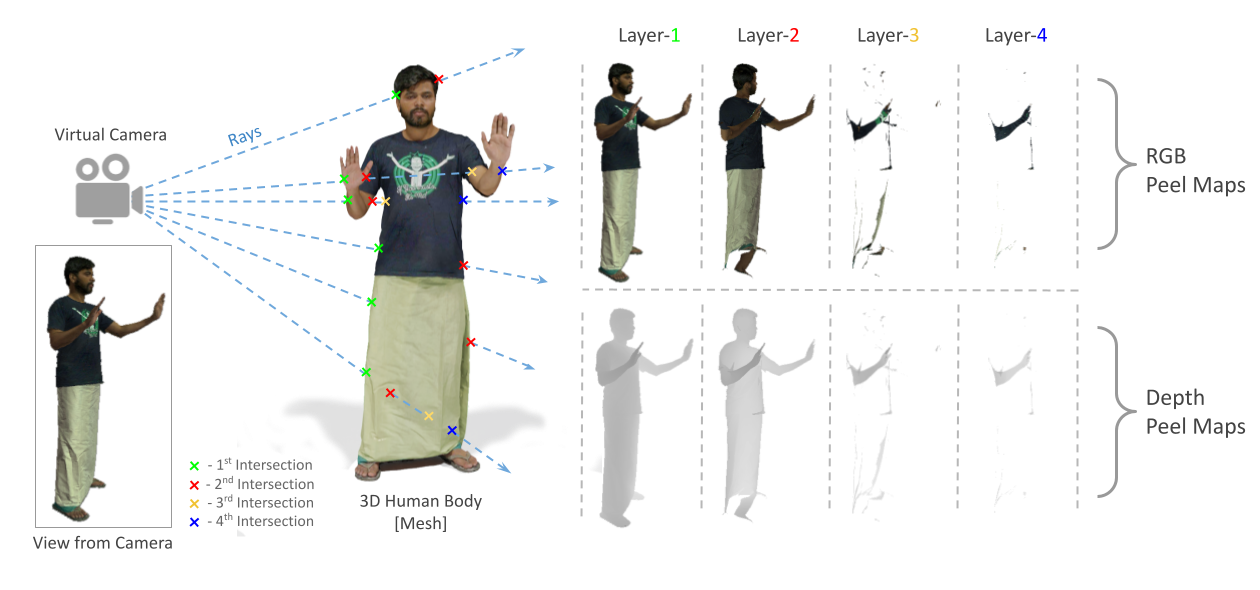}
            \captionsetup{justification=centering}
            \caption{PeeledHuman representation to encode 3D human body into 2D maps.}
            \label{fig:peeled_representation}
\end{figure*}

\section{Method}
    In this section, we first outline PeeledHuman representation for encoding 3D shapes and discuss briefly about SMPL, followed by the details of our proposed framework.
    \subsection{Background}
    \subsubsection{PeeledHuman Representation}\label{sec:PH-representation}
Our PeeledHuman representation is a sparse, non-parametric, multi-layered encoding of 3D shapes \citep{jinka2020peeledhuman}. The human body mesh is placed in a virtual scene and a set of rays are emanated from the camera center through each pixel towards the mesh.
The first set of ray-intersections with the mesh are recorded as the first layer depth peel map $d_1$ and RGB peel map $r_1$, capturing visible surface details nearest to the camera. This is similar to RGBD images captured from sensors, like Kinect. Subsequently, the rays are extended beyond the first intersection point (piercing through the intersecting surface) to hit the surface behind it. The corresponding depth and RGB values are recorded in the next layer peel maps, represented by $d_i$ and $r_i$ respectively, as shown in \autoref{fig:peeled_representation}. We use total $i=4$ layers of peeled representation in this work.
This representation is efficient, as it only stores ray-surface intersection in the form of sparse 2D maps, unlike voxels and implicit function representations, which are redundant in their representation.

\subsubsection{SMPL Parametric Body Model}
Skinned Multi-Person Linear (SMPL) \citep{SMPL:2015} is a parametric 3D model of the human body that is based on vertex-based skinning and blend shapes and is learned from thousands of 3D body scans. SMPL factors the full human body mesh into the pose ($\theta \in \mathbb{R}^{72}$) and shape ($\beta \in \mathbb{R}^{10} $)  parameters. $\theta$ for each joint is defined as the axis angle rotation relative to its parent in the kinematic tree, while $\beta$ represents the shape PCA coefficients learned from various body scans. 

    SMPL starts with an artist-created mean template mesh $\mathcal{T} \in \mathbb{R}^{6890 \times 3}$ and blend skinning weights $\mathcal{W} \in \mathbb{R}^{6890 \times 24}$. Template mesh, based on its skeleton joints, $\mathcal{J}(\cdot)$ is deformed through two blend functions $\mathcal{B}_s(\beta)$ and $\mathcal{B}_p(\theta)$. Shape blend-shape function $\mathcal{B}_s(\beta)$ performs the per-vertex displacements, sculpting the person's identity, whereas pose-dependent blend-shape function $\mathcal{B}_p(\theta)$ takes a vector of pose parameters $\theta$ as input and maps them to another set of additive per-vertex displacements. Pose-dependent blend-shape function accounts for dynamic soft tissue deformation caused by the pose deviation from the rest-pose. 
    Finally, the deformed template mesh $\mathcal{T}(\theta + \mathcal{B}_s(\beta) + \mathcal{B}_p(\theta))$ is transferred to the final mesh $\mathcal{M}(\theta, \beta)$ through a linear blend skinning (LBS) function $W(\cdot)$ as:
\begin{equation}
    \mathcal{M}(\theta, \beta) = W(\mathcal{T}(\theta + \mathcal{B}_s(\beta) + \mathcal{B}_p(\theta), \mathcal{J}(\beta), \mathcal{W})
\end{equation}

\begin{figure*}[!ht] 
\includegraphics[width=\linewidth]{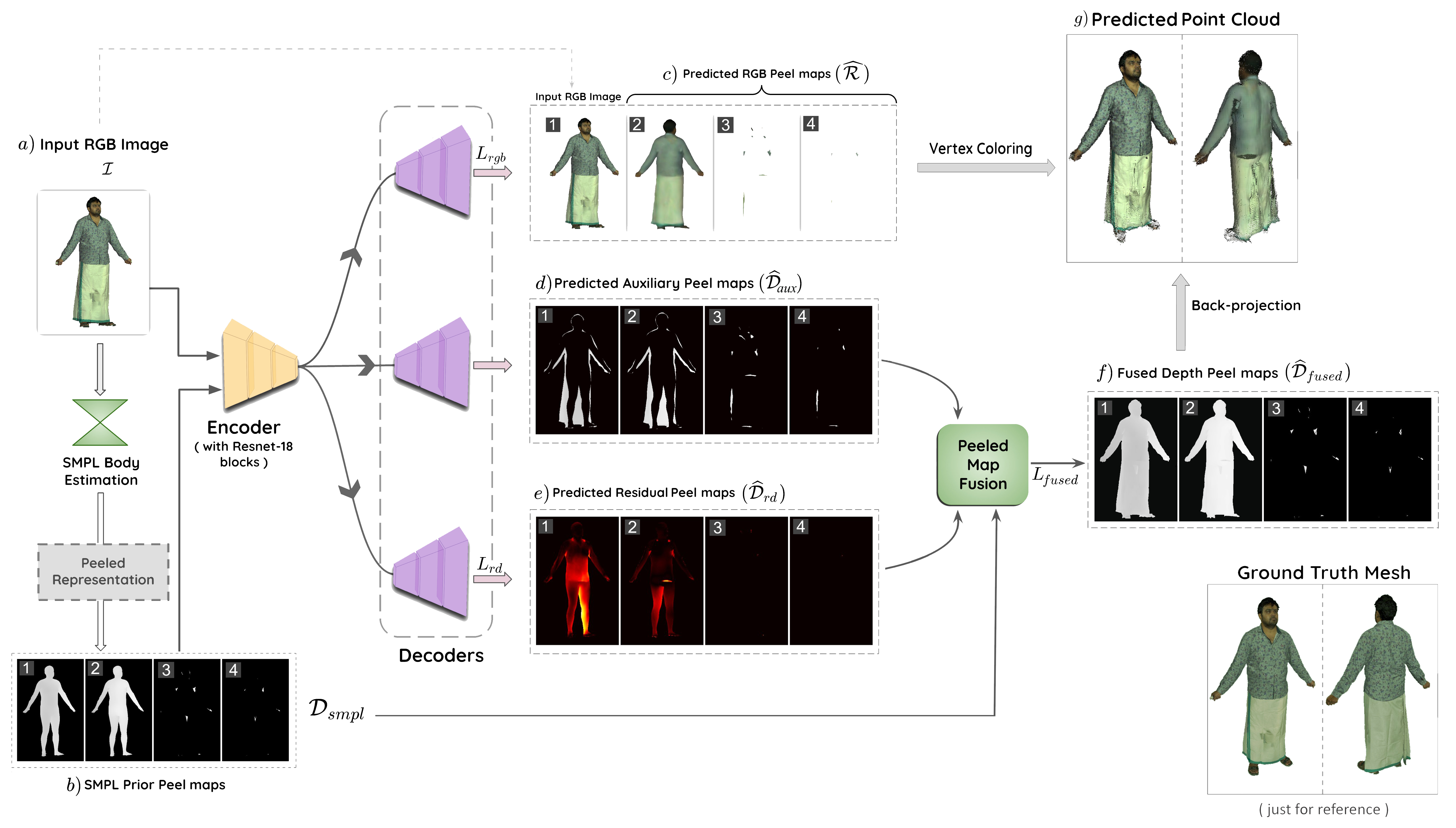}
\caption{\textbf{Pipeline:} We use an off-the-shelf method to estimate SMPL prior from the input image $\mathcal{I}$, and encode it into peeled representation ($\mathcal{D}_{smpl}$). This, along with image $\mathcal{I}$, is fed to an encoder. Subsequently, three separate decoders branches predict RGB peel maps ($\widehat{\mathcal{R}}$), auxiliary peel maps ($\widehat{\mathcal{D}}_{aux}$) and residual peel maps ($\widehat{\mathcal{D}}_{rd}$), respectively. Finally, a layer-wise fusion of $\widehat{\mathcal{D}}_{aux}$, $\widehat{\mathcal{D}}_{rd}$ and $\mathcal{D}_{smpl}$ is performed to obtain fused peel maps $\widehat{\mathcal{D}}_{fused}$, which is then back-projected along with $\widehat{\mathcal{R}}$ to obtain a vertex colored point-cloud. (The ground truth mesh is shown for comparison only.)} 
    \label{fig:Pipeline}
    \end{figure*}

\subsection{Overview} 
\label{subsec:peeled++}
We aim to reconstruct a 3D textured human body model of a person in arbitrary pose and clothing from a given monocular input image $\mathcal{I}$, as shown in \autoref{fig:Pipeline}.

Here, we discuss the steps involved in our proposed method.
\begin{enumerate}
    \item SMPL shape and pose parameters (i.e., $\beta \in \mathbb{R}^{10}$, $\theta \in \mathbb {R}^{72}$) along with parameters of weak perspective camera ($s, t_x, t_y$) are estimated from ProHMR \citep{kolotouros2021probabilistic}. We convert the estimated SMPL to depth peel maps which acts as a shape prior $\mathcal{D}_{smpl}$ (\autoref{fig:Pipeline}) as outlined in \autoref{subsubsec:peeled_shape_prior.}.
    \item Later, input image $\mathcal{I}$ (with background removed) is concatenated with $\mathcal{D}_{smpl}$ and is fed as an input to the shared encoder in our network. 
    \item Subsequently, three decoders predict different outputs through separate branches, namely, RGB peel maps $\widehat{\mathcal{R}}$, auxiliary peel maps $\widehat{\mathcal{D}}_{aux}$ and residual peel maps $\widehat{\mathcal{D}}_{rd}$, as shown in \autoref{fig:Pipeline} (c)-(e). 
The topmost decoder branch predicts only three RGB peel maps as the input $\mathcal{I}$ naturally acts as the first RGB peel map. 
    \item The SMPL prior peel maps $\mathcal{D}_{smpl}$, residual peel maps $\widehat{\mathcal{D}}_{rd}$ and auxiliary peel maps $\widehat{\mathcal{D}}_{aux}$ are further combined using SMPL mask $\Gamma_i$ (estimated using \autoref{eq:SMPL-mask}) to obtain the final fused peel maps $\widehat{\mathcal{D}}_{fused}$. 
    
    \item Finally, a colored point-cloud is obtained by back-projecting $\widehat{\mathcal{D}}_{fused}$ and $\widehat{\mathcal{R}}$ to camera coordinate system, as shown in \autoref{fig:Pipeline} (g). This point-cloud is further post-processed, and then meshified using Poisson Surface Reconstruction (PSR) \citep{kazhdan2006poisson}.
\end{enumerate}

To illustrate the importance of a shape prior in the prediction of peel maps, we compare SHARP with peeledhuman \citep{jinka2020peeledhuman}. The peeledhuman network predicts inconsistent body parts as shown in \autoref{fig:peeledhuman-errors} (a). This is because of the fact that there are no geometrical constraints imposed on the structure of predicted body parts.
The introduction of prior enables SHARP to reconstruct the human body with plausible body parts and accurate pose as shown in \autoref{fig:peeledhuman-errors} (b).

\subsection{Pipeline Details}
Here, we discuss in detail about our pipeline, which involves peeled shape prior, residual \& auxiliary peel maps and finally, peel map fusion. We also explain in detail the loss functions used to train SHARP.
\begin{figure*}[!]
\centering
\includegraphics[width=\linewidth]{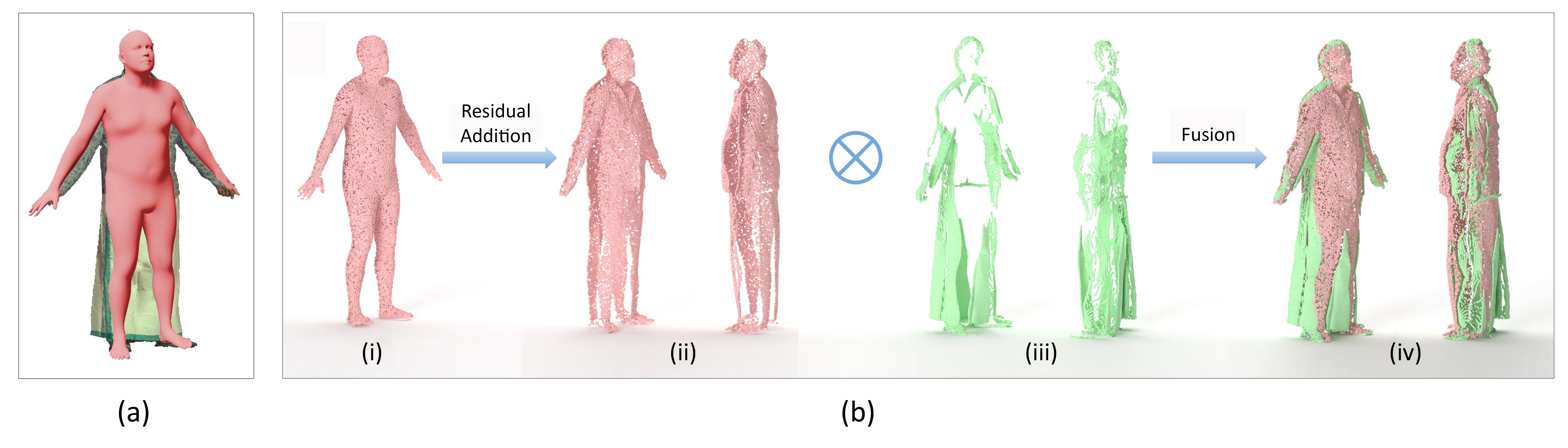}
\caption{\textbf{(a)} \textbf{SMPL prior overlayed on the input image}: The residual peel maps recover depth along the pixels over which SMPL prior is present across all the layers. For the remaining pixels, auxiliary peel maps are used to recover depth. \textbf{(b)} \textbf{3D representation of fusion:} (i) Point cloud obtained $\mathcal{D}_{smpl}$ is shown in red. (ii) Point cloud obtained from ($\mathcal{D}_{smpl}$ + $\widehat{\mathcal{D}}_{rd}$) is shown from two views in red. (iii) Point cloud obtained from $\widehat{\mathcal{D}}_{aux}$ is shown from two views in green. (iv) Final point cloud obtained from  $\widehat{\mathcal{D}}_{fused}$.}
            \label{fig:RD}
\end{figure*}

\subsubsection{Peeled Shape (SMPL) Prior} 
\label{subsubsec:peeled_shape_prior.}

We initially use \citep{kolotouros2021probabilistic} to estimate the SMPL pose and shape parameters ($\beta$, $\theta$), along with weak-perspective camera parameters ($s,t_x, t_y$).
The SMPL mesh is brought into the camera coordinate system using ($s,t_x, t_y$), and
then encoded into depth peel maps by passing camera rays through each pixel, as explained in \autoref{sec:PH-representation}, i.e., for every pixel $p$ in layer $i$, depth value of the point intersected by the camera ray is stored: 
        \begin{equation}
        \mathcal{D}_{smpl} = \{(d^i_{p}) : \forall p \in \mathcal{I}, i \in \{1,2,3,4\},d \in \mathbb{R}\}
        \end{equation}
We initialize a layer-wise binary SMPL mask $\Gamma_i$ by applying thresholding on SMPL prior peel maps. Additionally, we condition this mask on a pre-estimated binary foreground mask $\mathcal{F}$.
The foreground mask $\mathcal{F}$ covers only the clothed human in the input image and can be obtained using off-the-shelf background segmentation methods e.g. PGN \citep{gong2018instancelevel} .
We use $\mathcal{D}_{smpl}^i$ and $\mathcal{F}$ to estimate the per-layer SMPL mask $\Gamma_i$ as :
      \begin{equation}
            \Gamma_i= 
        \begin{cases}
            1,& \text{if } {\mathcal{D}_{smpl}^i}  \odot \mathcal{F} > 0 \text{ and } \\
            0,  & \text{otherwise.}
        \end{cases}
        \label{eq:SMPL-mask}
        \end{equation}
In essence, $\Gamma_i$ for each layer is estimated by retaining only the overlapping regions in corresponding SMPL prior peel map and the foreground mask. This helps refine the initial SMPL mask $\Gamma_i$ by eliminating parts of the SMPL prior peel maps that falls outside the human body \& clothing silhouette, thereby enabling our method to partially overcome the misalignment of SMPL prior with the input image. Note that, $\mathcal{F}$ is common across all the layers. The refined SMPL mask $\Gamma_i$ is subsequently used for peel map fusion in \autoref{eq:fusion}.  

    \begin{figure}
            \centering
            \includegraphics[width= 0.9\linewidth]{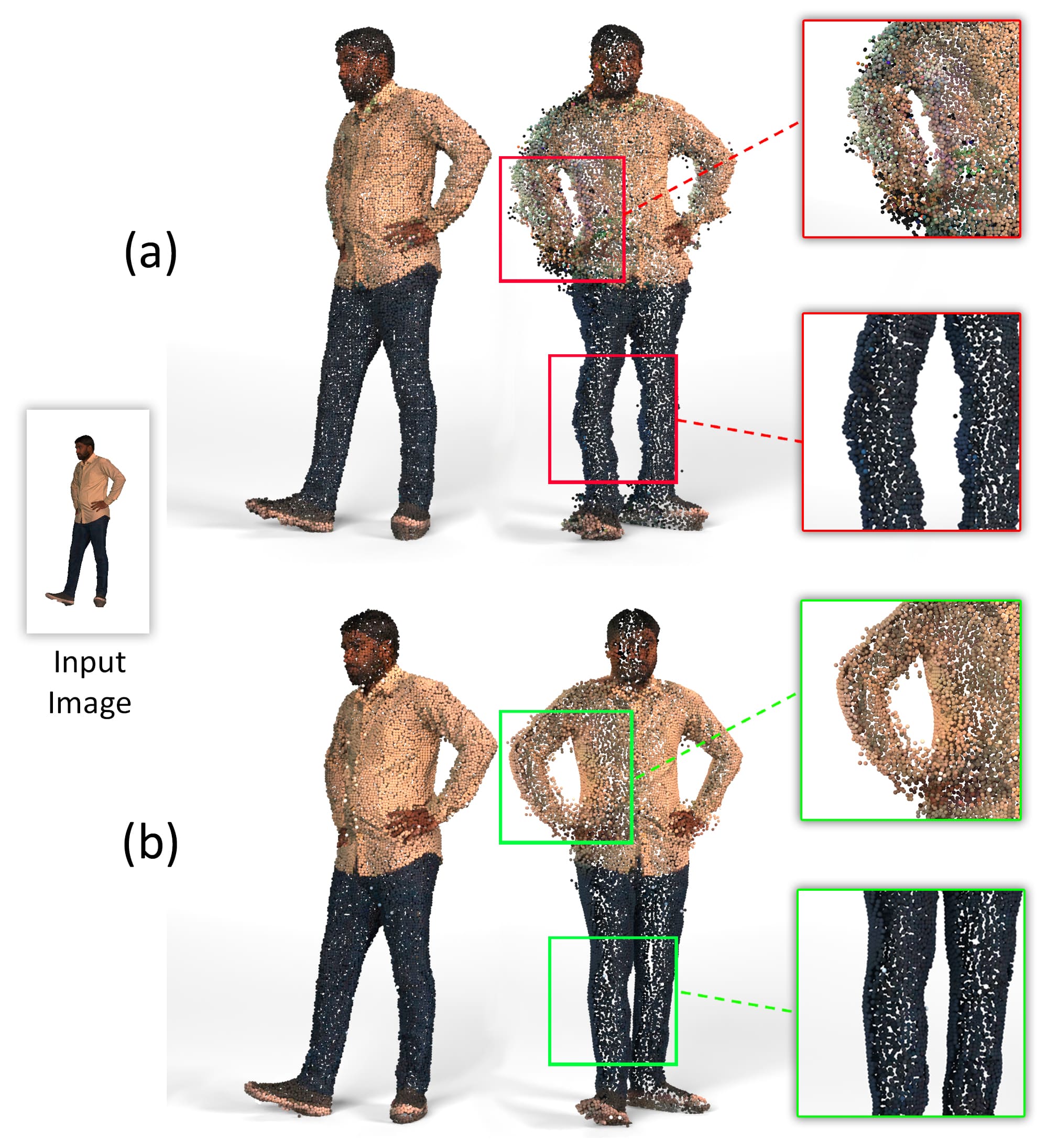} 
            \caption{(a) Distorted  body parts in the prediction from peeledhuman \citep{jinka2020peeledhuman}. (b) Reconstruction obtained from SHARP.}
            \label{fig:peeledhuman-errors}
\end{figure}        

\subsubsection{Residual and Auxiliary Peel Maps} 
\label{subsubsec:RD}
To estimate view specific deformations from the SMPL prior input, we propose to predict residual peel maps $\widehat{\mathcal{D}}_{rd}$ by computing additive pixel-wise offsets from the input SMPL depth peel maps $\mathcal{D}_{smpl}$. For every pixel $p$ in layer $i$ of peeled SMPL prior, we predict offset along z-axis \footnote{The camera is placed at  (0, 0, 10), Y axis is up and -Z axis is forward, while meshes are placed at origin.}: 
    \begin{equation}
        \widehat{\mathcal{D}}_{rd} = \{(\widehat{\delta}^i_{p}) : \forall p \in \mathcal{I}, i \in \{1,2,3,4\}, \widehat{\delta} \in \mathbb{R}\}
    \end{equation}
    
For pixels, which depict the projection of bare body parts, 
network predicts minimal offsets ($\widehat{\mathcal{D}}_{rd}$), thereby capturing the person-specific appearance features like hairline and facial details while preserving overall structure of the body parts.

Thus, each layer of the residual peel maps provides pixel-wise displacements of the corresponding layer of peeled SMPL prior maps along the view-direction (z-axis). These residual deformations only cover the pixels in the input image for which SMPL prior is present. For the remaining pixels of clothed body, we propose to learn their depth values using a separate branch in the form of auxiliary peel maps.
\begin{equation}
        \widehat{\mathcal{D}}_{aux} = \{(\widehat{d}^i_{aux}) : \forall p \in \mathcal{I}, i \in \{1,2,3,4\}, \widehat{d} \in \mathbb{R}\}
\end{equation}

\autoref{fig:RD} provide 3D visualization of ($\mathcal{D}_{smpl}$ + $\widehat{\mathcal{D}}_{rd}$) and
$\widehat{\mathcal{D}}_{aux}$ by back-projecting respective partial depth peel maps. We can observe that these two capture the complementary geometrical details of 3D body and clothing. 
\subsubsection{Peel Map Fusion}\label{subsubsec:fusion}

The predicted residual and auxiliary peel maps independently capture complimentary surface details and are subsequently fused to obtain the geometry of unified clothed body.
We propose to obtain final fused peel depth maps by layer-wise fusion of $\mathcal{D}_{smpl}$, $\widehat{\mathcal{D}}_{rd}$ and $\widehat{\mathcal{D}}_{aux}$ expressed as:

\begin{equation}
    \widehat{\mathcal{D}}_{fused} = ({\mathcal{D}}_{smpl}+\widehat{\mathcal{D}}_{rd}) \otimes \widehat{\mathcal{D}}_{aux}
\label{eq:fusion_eq}
\end{equation}
where $\otimes$ is the proposed layer-wise fusion operator as explained below.
Here,
        \begin{equation}
            \widehat{\mathcal{D}}_{fused}^i = \Gamma_{i} \odot (\widehat{\mathcal{D}}_{rd}^i+\mathcal{D}_{smpl}^i) + (1-\Gamma_{i}) \odot \widehat{\mathcal{D}}_{aux}^{i}
        \end{equation} \label{eq:fusion} 
        here, $\odot$ is element-wise multiplication and for each $i^{th}$ layer $\widehat{\mathcal{D}}_{aux}^i \in \widehat{\mathcal{D}}_{aux}$, $\widehat{\mathcal{D}}_{rd}^i \in \widehat{\mathcal{D}}_{rd}$ and $\mathcal{D}_{smpl}^i \in \mathcal{D}_{smpl}$. 

In summary, we have decoupled the task of recovering the clothed 3D human body surface into predicting residual and auxiliary peel maps. We later fused these partial reconstructions into a single unified 3D surface. 
Our approach ensures geometrically consistent body parts as the residual peel maps predict minimal offsets on the pixels belonging to the bare body where there is no clothing, thereby retaining body-specific geometry.

\subsection{Loss Functions}\label{subsec:losses}
We use encoder-decoder architecture for our predictions in SHARP. We train our network with losses on 2D peel map predictions.
Our final learning objective is defined as:
        \begin{equation}
            L = L_{fuse} + \lambda_{rd} L_{rd} + \lambda_{rgb} L_{rgb} + \lambda_{sm} L_{sm}
        \end{equation}
where $\lambda_{rd}$, $\lambda_{rgb}$ and $\lambda_{sm}$ are regularization parameters for $L_{rd}, L_{rgb}, L_{sm}$, respectively. We provide the formulation for the individual loss terms below.

\begin{equation}
            L_{fuse} = \sum_{i=1}^{4} \Big\lVert {\widehat{ \mathcal{D}}_{fused}^i - \mathcal{D}_{fused}^i}\Big\lVert_1
\end{equation}

$L_{fuse}$ is the sum of $L_1$ norm between ground truth depth peel maps $\mathcal{D}_{fused}$ and predicted fused peel maps $\widehat{\mathcal{D}}_{fused}$ for each $i^{th}$ layer.

\begin{equation}
            L_{rd} = \sum_{i=1}^{4} \Big\lVert{\widehat{ \mathcal{D}}_{rd}^i - \mathcal{D}_{rd}^i}\Big\lVert_1
\end{equation}

$L_{rd}$ constraints the residual peel map predictions to that of ground truth offsets. Note that we are training auxiliary peel maps branch without any explicit loss on $\widehat{\mathcal{D}}_{aux}$. The gradients through auxiliary peel map branch back-propagates using $L_{rd}$ and $L_{fuse}$.

We also enforce per layer first order gradient smoothness of the predicted $(\widehat{\mathcal{D}}_{rd}^i+\mathcal{D}_{smpl}^i)$ and ground truth  $({\mathcal{D}}_{rd}^i+\mathcal{D}_{smpl}^i)$ as well as between ground truth and predicted $\widehat{\mathcal{D}}_{fused}$ maps. $L_{sm}^{fuse}$ ensures smoothness between the two predicted surfaces.
\begin{equation}
    L_{sm} = L_{sm}^{rd} + L_{sm}^{fuse}
\end{equation}
where,

\begin{equation}
                \begin{aligned}
                    L_{{sm}}^{rd} &= \sum_{i=1}^{4} \Big\lVert{\bigtriangledown ({\mathcal{D}}_{rd}^i+\mathcal{D}_{smpl}^i) - \bigtriangledown (\widehat{\mathcal{D}}_{rd}^i+\mathcal{D}_{smpl}^i)}\Big\Vert_1 \\
                    L_{{sm}}^{fuse} &= \sum_{i=1}^{4} \Big\lVert{\bigtriangledown {\mathcal{D}}_{fused}^i - \bigtriangledown \widehat{\mathcal{D}}_{fused}^i}\Big\Vert_1 
                \end{aligned}
            \label{eq:smoothloss}
\end{equation}
Additionally, We also train our network with $L_1$ loss between predicted and ground truth RGB peel maps ($L_{{rgb}}$).

\section{3DHumans Dataset}
\label{sec:our_data}
\begin{figure*}[h!]
\centering
        \includegraphics[width=0.8\linewidth]{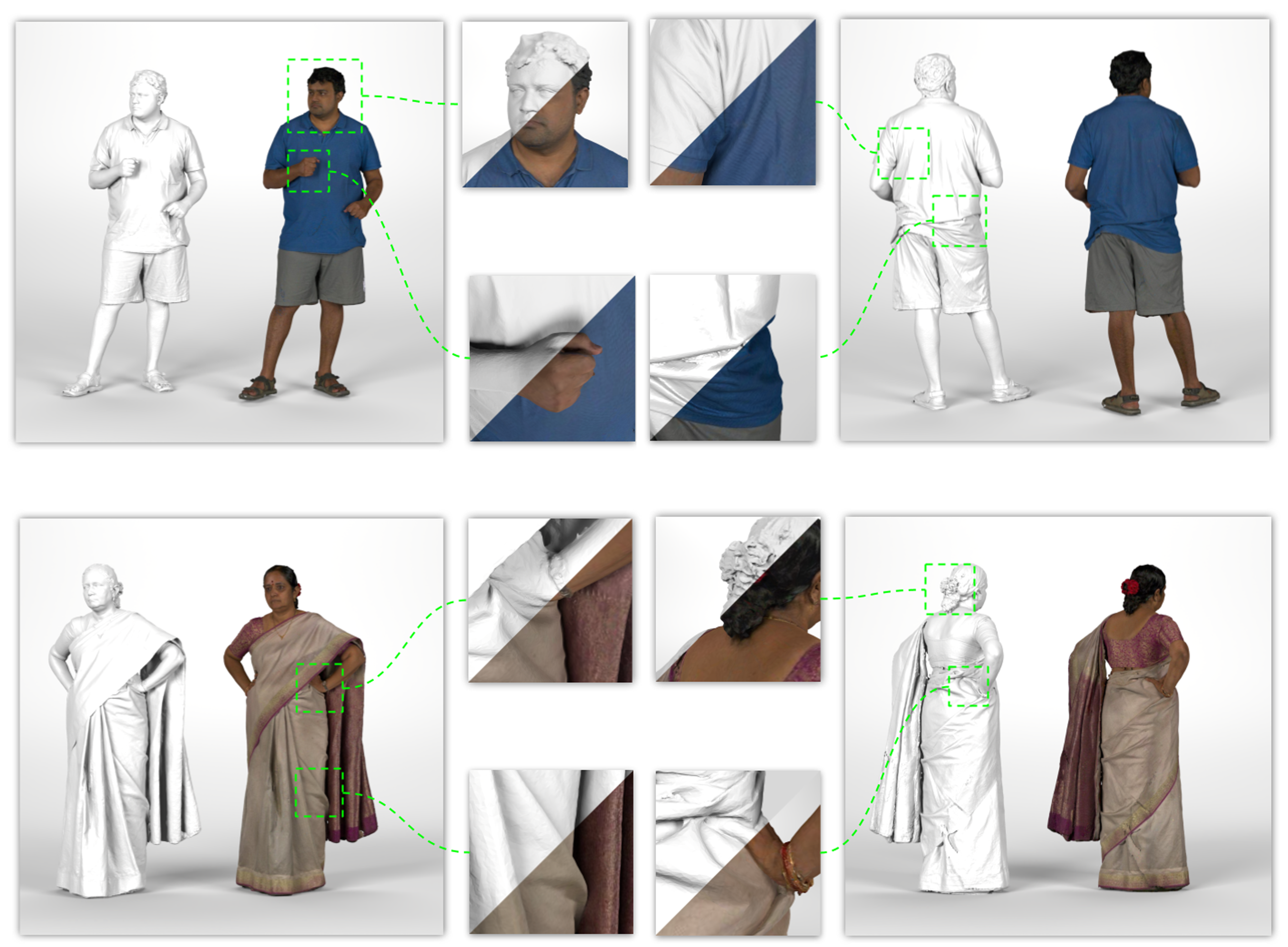}
        \caption{High-frequency geometrical and textural details present in our 3DHumans dataset. }
        \label{fig:our_dataset}
    \end{figure*}
As mentioned in \autoref{sec:introduction}, one of the key bottlenecks that hinder progress in the field of 3D human body reconstruction is the lack of publically available real-world datasets that contain high-frequency texture and geometrical details. 

To this end, we present 3DHumans, a dataset of around 250 scans containing people in diverse body shapes in various garments styles and sizes.
We cover a wide variety of clothing styles ranging from loose robed clothing like saree (a typical South-Asian dress) to relatively tight-fitting shirt and trousers, as shown in \autoref{fig:our_dataset}. 
The dataset consists of around 150 male and 50 unique female subjects. Total male scans are about 180 and female scans are around 70. 
In terms of regional diversity, for the first time, we capture body shape, appearance and clothing styles for the South-Asian population. We will release this data in the public domain for academic use.\footnote{\href{http://cvit.iiit.ac.in/research/projects/cvit-projects/sharp-3dhumans-a-rich-3d-dataset-of-scanned-humans}{http://cvit.iiit.ac.in/research/projects/cvit-projects/sharp-3dhumans-a-rich-3d-dataset-of-scanned-humans}}

The 3DHumans dataset is created using the Artec3D Eva hand-held structured light scanner. The scanner has a 3D point accuracy of up to 0.1mm and 3D resolution is 0.5mm. For each 3D human scan, we also provide the SMPL body aligned to it, using \citep{zheng2021pamir,pavlakos2019expressive}.

\section{Experiments \& Results}
In this section, we present the experimental details, datasets and training protocol for SHARP. We also show qualitative and quantitative comparisons with current state-of-the-art methods.
\subsection{Implementation Details}
We employ a multi-branch encoder-decoder network for SHARP, which is trained in an end-to-end fashion. The network takes the input image concatenated with SMPL peel maps in $512 \times 512$ resolution. The shared encoder is consist of a convolutional layer and $2$ downsampling layers which have $64, 128, 256$ kernels of size 7$\times$7, 3$\times$3 and 3$\times$3, respectively. This is followed by ResNet blocks which take downsampled feature maps of size 128$\times$128$\times$256.
The decoders for predicting $\widehat{\mathcal{D}}_{fused}$, $\widehat{\mathcal{D}}_{rd}$ and $\widehat{\mathcal{R}}$, consist of two upsampling layers followed by a convolutional layer, having same kernel sizes as of the shared encoder.
Sigmoid activation is used in $\widehat{\mathcal{D}}_{fused}$ and $\widehat{\mathcal{D}}_{rd}$ decoder branches, while a tanh activation is used for the $\widehat{\mathcal{R}}$ decoder branch. The $\widehat{\mathcal{D}}_{rd}$ output values are scaled to a $[-1,0.5]$ range which is found empirically.

We use the Adam optimizer with an exponentially decaying learning rate starting from $5 \times 10^{-4}$. Our network takes around 18 hrs to train for 20 epochs on $4$ Nvidia GTX $1080$Ti GPUs with a batch size of $8$ and $\lambda_{rd}$, $\lambda_{fuse}$, $\lambda_{rgb}$ and $\lambda_{sm}$ are set to $1, 1, 0.1$ and $0.001$, respectively, found empirically. We use trimesh \citep{trimesh} library for rendering the peel maps.

\subsection{Other Datasets}\label{sec:other-datasets}
In addition to our 3DHumans dataset (\autoref{sec:our_data}), we perform both qualitative and quantitative evaluations on the following publicly available datasets.
\\
\\
\textbf{CLOTH3D} \citep{bertiche2020Cloth3D} is a collection of $6500$ synthetic sequences of SMPL meshes with garments draped onto them, simulated with MoCap data. Each frame of a sequence contains garment and corresponding SMPL body. The garment styles range from skirts to very loose robes. 
We augment this data by capturing SMPL texture maps with minimal clothing to simulate realistic body textures using  \citep{alldieck19cvpr}. For each sequence, five frames are randomly sampled. Please refer to the supplementary draft \textbf{(section 1)} for understanding the data preparation step and results of SHARP on CLOTH3D.\\
\\
\textbf{THUman1.0} \citep{zheng2019deephuman} consists of $6800$ human meshes registered with SMPL body in varying poses and garments. The dataset was obtained using consumer RGBD sensors. Although the dataset has diverse poses and shapes, it has relatively tight clothing examples with low-quality textures. Please refer supplementary for results on this dataset. Note that the dataset is originally called the \textbf{THUman} dataset, we refer it to as \textbf{THUman1.0} to avoid the confusion.\\
\\
\textbf{THUman2.0} \citep{yu2021function4d} is a collection of 500 high quality 3D scans captured using dense DSLR rig. 
The ataset offers wide variety of poses. However, very loose clothing styles like robed skirts are still lacking.
Each mesh in the provided dataset is in different scale. We have brought all the meshes in the same scale by registering SMPL to the scans and performed our experiments.

\begin{table*}

\makebox[\textwidth][c]{
\begin{tabular}{c|ccc|ccc}
\toprule
      &       & \multicolumn{1}{c|}{Our Dataset} & \multicolumn{2}{c}{THUman2.0 Dataset} \\
\midrule
Method & CD ($\times 10^{-5}$) $\downarrow$ & P2S $\downarrow$ & Normal $\downarrow$ & CD ($\times 10^{-5}$) $\downarrow$ & P2S $\downarrow$ & Normal $\downarrow$\\ \hline
PIFu & {20.79} & {0.00826} &{0.054}& {23.72} & {0.0091}&{0.036} \\
Geo-PIFu & {15.73} & {0.0092}&{0.058} & {17.01} &{0.0092}&{0.041} \\
PaMIR & {12.54} & {0.00714}&{0.054} & {6.05} & \textbf{0.0049}&{0.038} \\
PeeledHuman & {20.88} & {0.0094}&{0.061} & {23.34} & {0.0094}&{0.054}\\ \hline
Ours & \textbf{7.718} & \textbf{0.0051}&\textbf{0.045} & \textbf{6.044}&{0.00529}&\textbf{0.034}\\
\bottomrule
\end{tabular}%
}
\caption{Quantitative comparison on 3DHumans and THUman2.0 datasets.}
\label{tab:comparison_ours}
\end{table*}


\begin{figure*}[t!]
    \centering
        \includegraphics[width=0.9\linewidth]{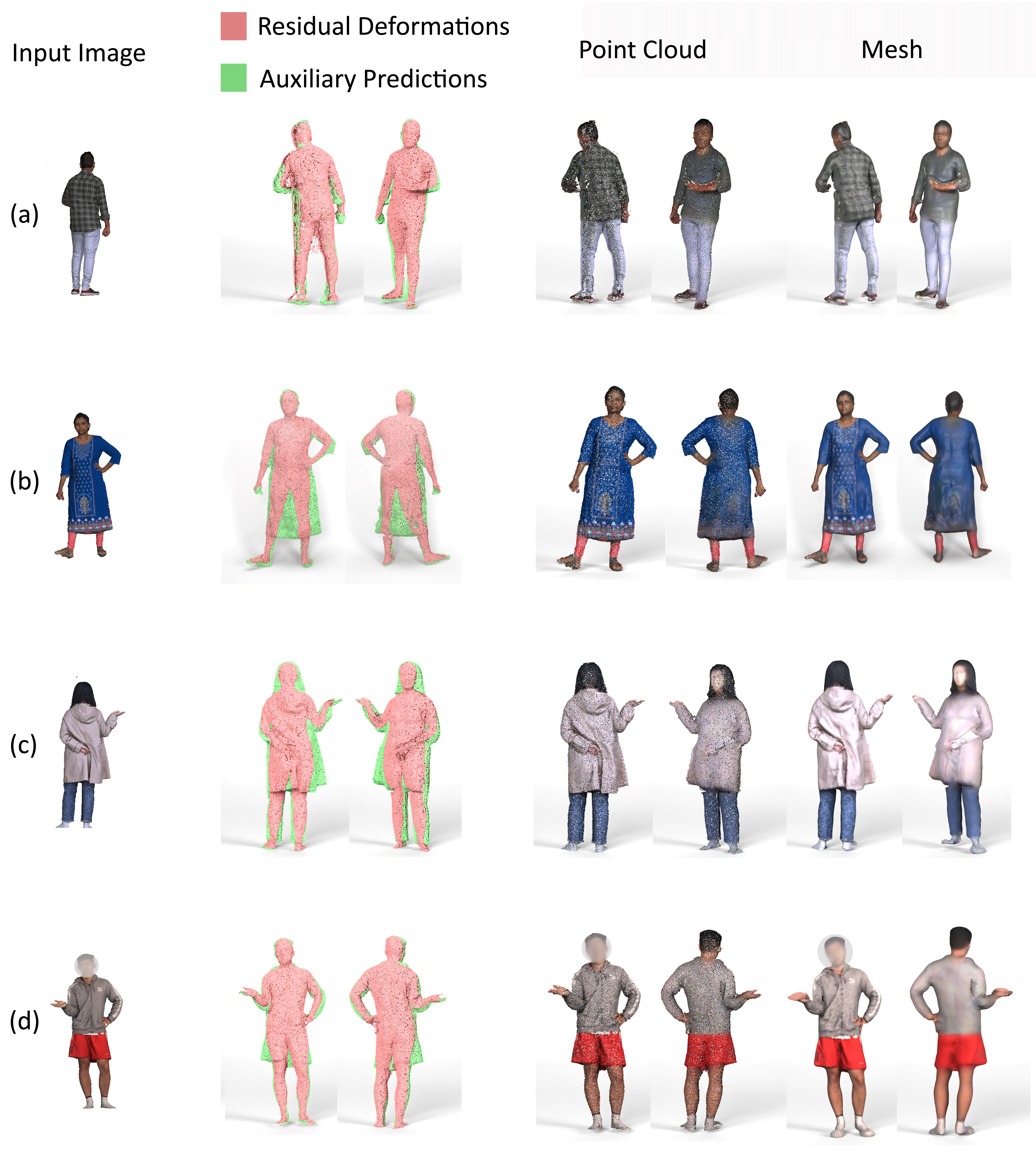}
        \caption{Results on 3DHumans (a, b) and THUman2.0 datasets (c, d).}
        \label{fig:results_our_dataset}
    \end{figure*}
  \begin{figure*}[t!]
        \includegraphics[width=\linewidth]{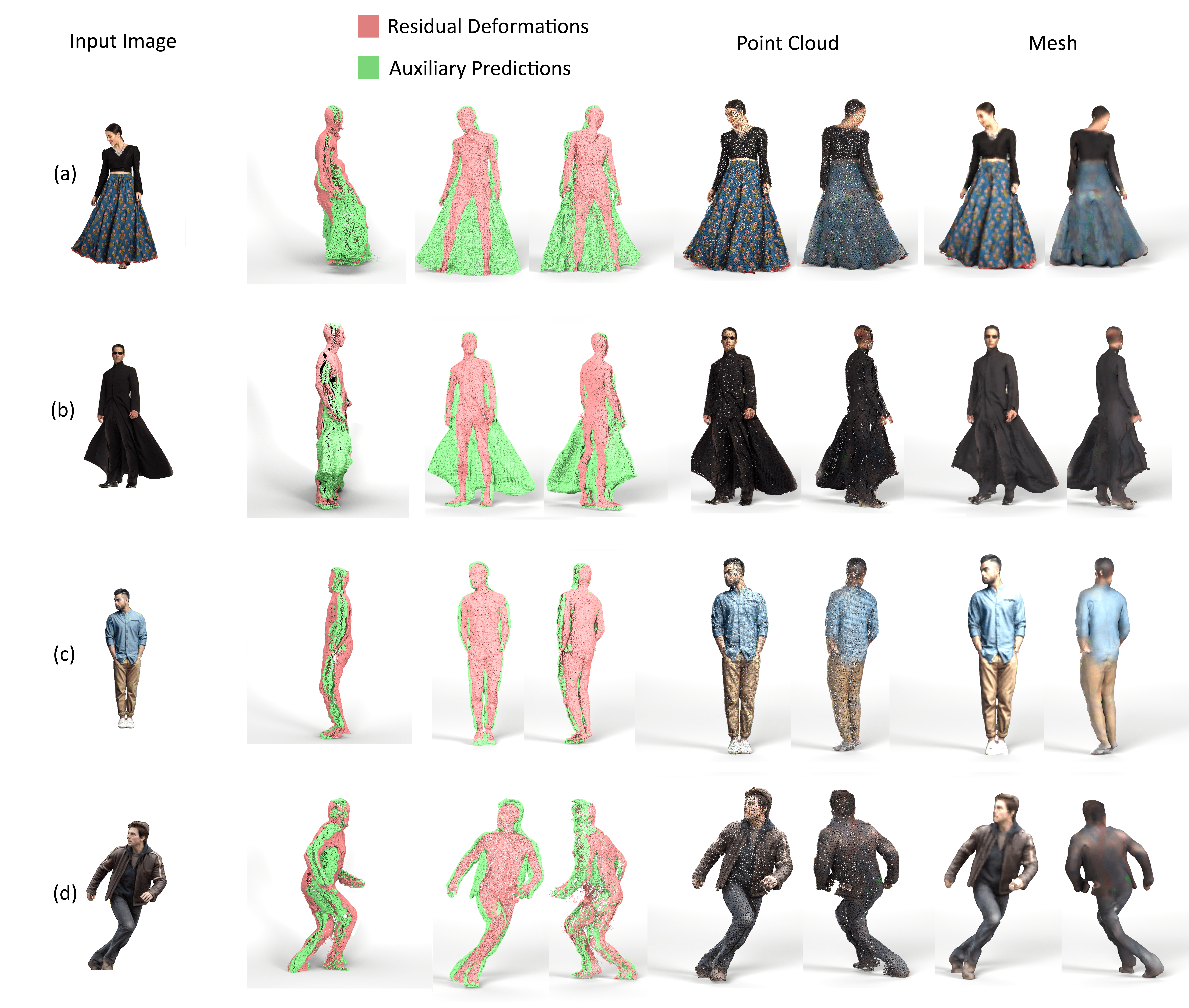}
        \caption{Results on in-the-wild images.}
        \label{fig:results_real_images}
    \end{figure*}

\begin{figure*}
        \hspace*{0.5in}
        \includegraphics[width=0.8\linewidth]{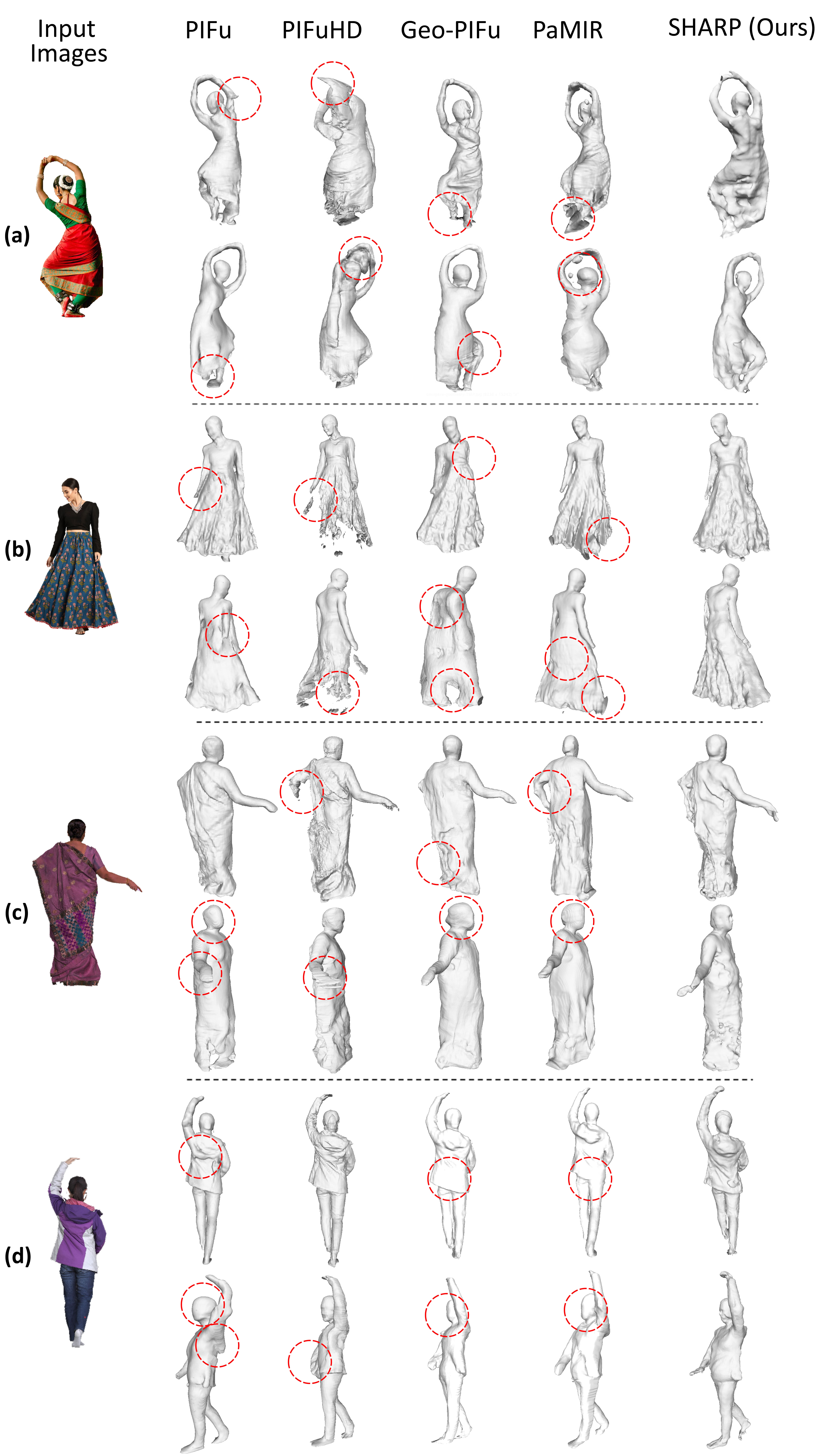}
        \caption{Qualitative comparison of SOTA methods. (a) and (b) are in-the-wild images, (c) and (d) are from 3DHumans and THUman2.0 datasets respectively, shown in two different views.}
        \label{fig:results_compare_dataset}
    \end{figure*}
    \begin{figure}
        \includegraphics[width=\linewidth]{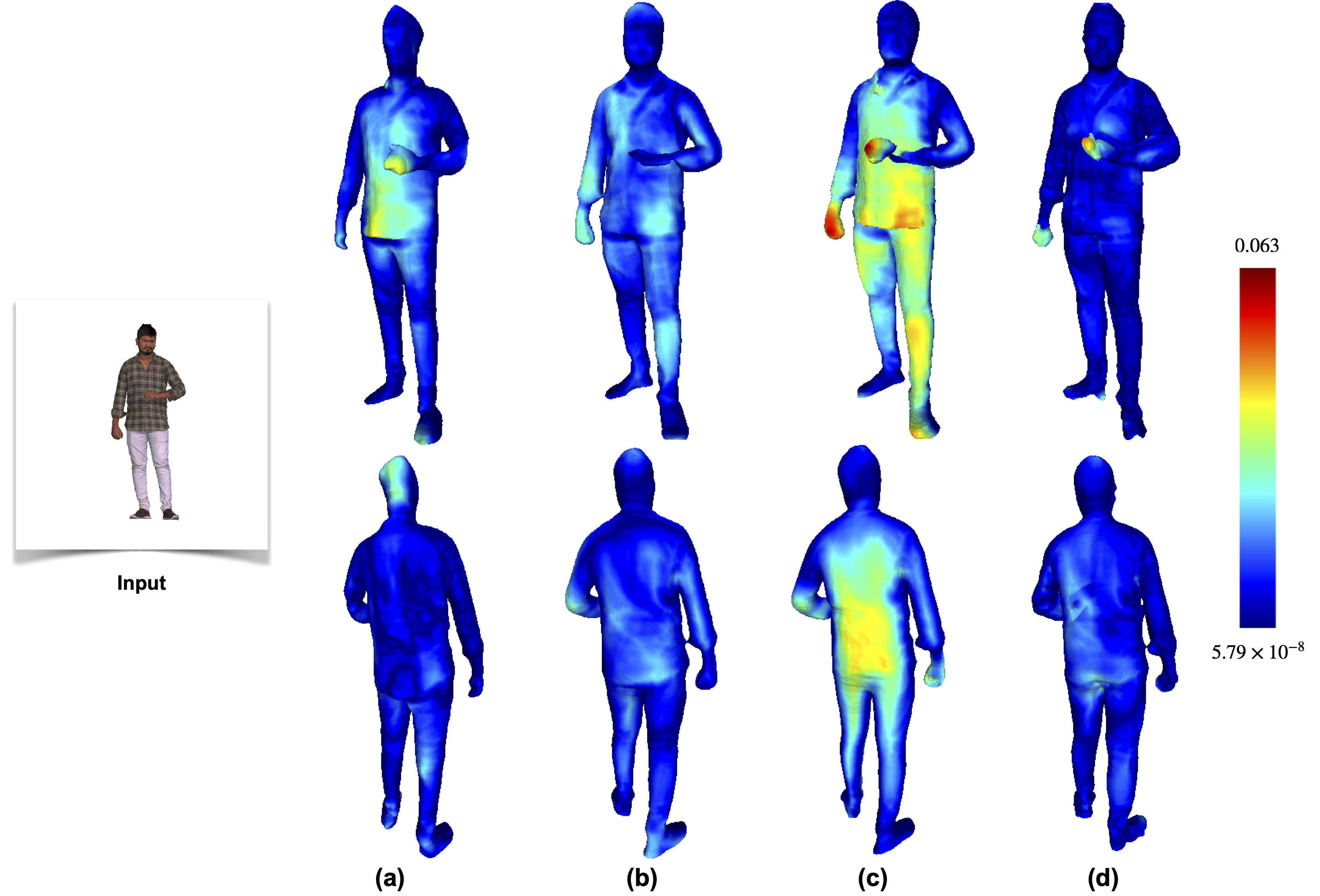}
        \caption{\textbf{P2S Plot:} Point-to-surface plots on the reconstructed outputs from (a) PaMIR, (b) Geo-PIFu, (c) PIFu and (d) SHARP.}
        \label{fig:p2s}
    \end{figure}
       \begin{figure*}
        \includegraphics[width=\linewidth]{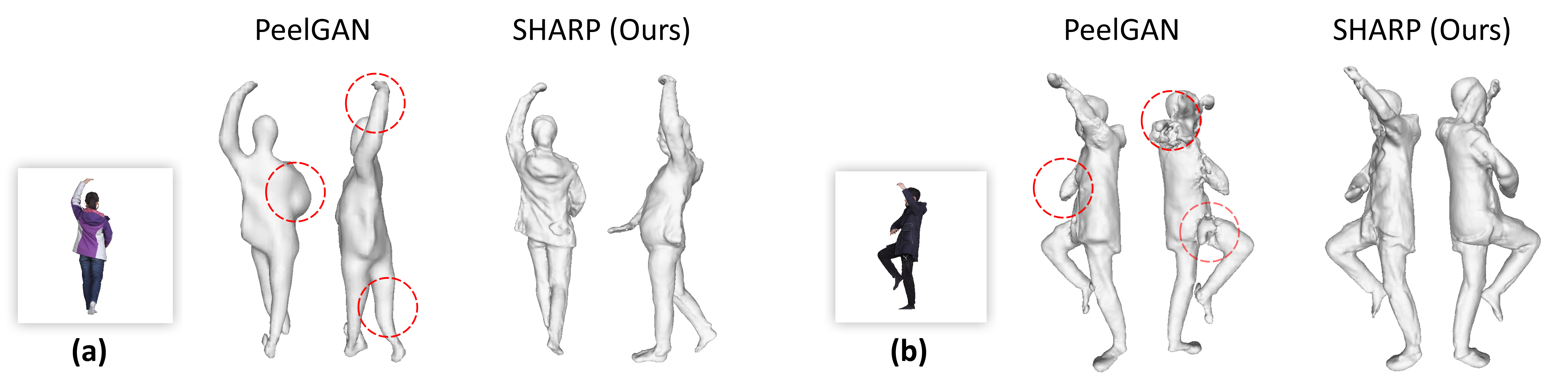}
        \caption{Qualitative comparison of peeledhuman and SHARP.}
        \label{fig:peeledhuman_compare}
    \end{figure*}

\begin{table}
            \centering
            \begin{tabular}{lcc}
            \toprule
            Method & CD $\downarrow$ & P2S $\downarrow$\\
            \midrule
                JumpSuit & 0.00031 & 0.00872\\
                Dress &  0.0012& 0.021\\
                Top+Trousers & 0.00057 & 0.0118\\
            \bottomrule
            \end{tabular}
            \caption{Performance of our method on clothing styles of CLOTH3D dataset.}
            \label{tab:dress_comparisons}
            \end{table}

\subsection{Evaluation Metrics}
To quantitatively evaluate performance of SHARP, we use the following evaluation metrics:  \\
\\
\textbf{Point-to-Surface (P2S) Distance:} Given a set of points and a surface, P2S measures the average L2 distance between each point and the nearest point to it on the given surface. We use P2S to measure the deviation of the point cloud (back-projected from predicted fused peel maps) from the ground truth mesh.
\linebreak
\linebreak
\textbf{Chamfer Distance (CD):} Given two sets of points $S_1$ and $S_2$, Chamfer distance measures the discrepancy between them as follows:

\begin{equation}
                \begin{aligned}
                    d_{CD}(S_1, S_2) &= \sum_{x \epsilon S_1} min_{y \epsilon S_2} \Vert x-y \Vert _2^2 \\
                    &+ \sum_{y \epsilon S_2} min_{x \epsilon S_1} \Vert x-y \Vert _2^2\\
                \end{aligned}
            \label{eq:smoothloss}
\end{equation}
\linebreak
\textbf{Normal Re-projection Loss:} To evaluate the fineness of reconstructed quality, we compute normal reprojection loss introduced in \citep{saito2019pifu}. We render the predicted and ground truth normal maps in the image space from the input viewpoint. We then calculate the L2 error between these two normal maps.

\subsection{Quantitative Evaluation}
We evaluate the aforementioned metrics on 3DHumans \& THUman2.0 datasets and comapred the results with PIFu  \citep{saito2019pifu}, PaMIR \citep{zheng2021pamir}, Geo-PIFu \citep{he2020geo} and PeeledHuman \citep{jinka2020peeledhuman}. We trained all the models from scratch on these datasets under the same train/test split. We transform all the predicted models from different methods to the canonical coordinates of the ground truth mesh and report metrices in \autoref{tab:comparison_ours}. The quantitative comparison concludes that our method outperforms the SOTA methods.

Unlike peeledhuman  \citep{jinka2020peeledhuman} that uses a generative network, we use a simple encoder-decoder architecture. We trained PaMIR with approximately thrice the amount of data (SHARP is trained on 70 views per mesh, while for PaMIR, 200 views per mesh are used). 
Geo-PIFu needs to be trained for coarse and query networks separately and complete training takes three days to train on 3DHumans in our setup.

Additionally, \autoref{tab:dress_comparisons} summarizes quantitative analysis on the CLOTH3D dataset where we evaluate CD and P2S metrics on different styles of clothing to indicate the generalization of our method across various clothing styles.
We also provide comparisons with THUman1.0 and CLOTH3D datasets in the supplementary. 

\subsection{Qualitative Evaluation}
We show the reconstructions obtained by our method using THUman2.0 and 3DHumans datasets (\autoref{fig:results_our_dataset}), where we also show point clouds obtained by back-projecting residual and auxiliary peel maps.    
\autoref{fig:results_our_dataset} (a) and (b) are samples from our dataset, (c) and (d) are from THUman2.0 dataset. Please refer supplementary for additional results on CLOTH3D dataset.
One can observe that our model can handle various styles of clothing (including \textit{tunic}) covering the lower body parts and with a wide variety of poses. 
Residual and auxiliary peel maps captured the complimentary surface details as visualized in red and green in \autoref{fig:results_our_dataset}.
In order to test the generalizability of our method on unseen in-the-wild images, 
we show results on random internet images using our method in \autoref{fig:results_real_images}. Similar to PIFu \citep{saito2019pifu}, we use an off-the-shelf method to remove the background from these images before passing them to our network.
It can be noted that our method is able to reconstruct the human body with self-occlusions and tackle a wide variety of clothing styles, ranging from tight to loose clothing with diverse poses. Notably, our method is able to generalize well on unseen, very loose clothing styles present in  \autoref{fig:results_real_images} (a) \& (b).

In \autoref{fig:results_compare_dataset}, we show qualitative comparison with SOTA methods. 
PIFu and PIFuHD do not use body prior, which leads to missing and distorted body parts. Geo-PIFu predicts a volumetric prior before performing implicit reconstruction.
On the other hand, PaMIR uses SMPL prior as input. Hence, both methods tends to produce smoother geometry as they use voxelized representation, which is known to smooth out the geometrical details.
It can be noted that our method retains high-frequency surface details as shown in \autoref{fig:results_compare_dataset}. 
Additionally, we also show comparison with our previous work peeledhuman \citep{jinka2020peeledhuman} in \autoref{fig:peeledhuman_compare}. We observed that our formulation yielded superior results over  peeledhuman which also uses the PeeledHuman representation sans SMPL prior. 

All the aforementioned methods have been trained on our 3DHumans dataset except for PIFuHD. Since the training code for PIFuHD is not yet available,  we use the model provided by the authors. 
In order to fairly compare with other methods, we selected a body with tight clothing and generated plots of P2S error of all methods trained on our dataset as visualized in \autoref{fig:p2s}. One can infer from these plots that our approach yields superior performance in terms of distribution of P2S error over the reconstructed surface.

\subsection{Network Complexity}
\begin{table*}[!h]
    \centering
\begin{tabular}{lcc}
\toprule
Method & No. of parameters & Execution Time \\
\midrule
PaMIR(Geo+Tex) & 40M(27M+13M) & 4.03s(3.9s+0.13s)\\
Geo-PIFu(coarse+fine) & 30.6M (14.9M+15.7M) & 16.32s(0.32s+16s)\\
Ours & \textbf{22M} & \textbf{0.09s}\\
\bottomrule
\end{tabular}

\caption{Comparison of complexity analysis.}
\label{tab:network_complexity}
\end{table*}
We report a detailed analysis of the execution time of SOTA methods in \autoref{tab:network_complexity}. 
All the numbers are computed on a single NVIDIA GTX 1080Ti GPU with a single input image.
PaMIR needs feed-forward of two networks to infer shape and geometry. On the other hand, Geo-PIFu needs to infer coarse volumetric shape followed by fine shape. We calculate the feed-forward execution time for the complete forward pass of Geo-PIFu and PaMIR as these methods need multiple forward passes while inferring. Note that ours is an end-to-end inference model which predicts both shape and color in a single forward pass efficiently with 0.09 seconds, which is significantly faster when compared to the aforementioned methods. 
Additionally, our network is lightweight, consisting of $22$ million parameters, while PaMIR and Geo-PIFu has $40$ and $30.6$ million parameters, respectively.

\begin{table}
      \centering
        \begin{tabular}{lcc}
            \toprule
            Method & CD $\downarrow$ & P2S $\downarrow$\\
            \midrule
                Ours w.o. $L_{sm}$ & 8.3652 & 0.0053\\
                Ours w.o. fusion &  9.98 & 0.0054\\
                Ours & \textbf{7.718} & \textbf{0.0051}\\
            \bottomrule
            \end{tabular}
            \caption{\textbf{Ablation Study:} Effect of loss functions.}
            \label{table:loss}
            
\end{table}

\begin{table}
      \centering
        \begin{tabular}{lccc}
            \toprule
            Network &  CD $\downarrow$ & P2S $\downarrow$\\
            \midrule
                {U-Net}  & {8.417} & {0.0052}\\
                {Hourglass} &{15.6} & {0.0068}\\
                {ResNet(ours)} & \textbf{7.71} & \textbf{0.0051}\\
            \bottomrule
        \end{tabular}
        \caption{\textbf{Ablation Study:} Effect of different architectures.}
        \label{tab:architectures}
\end{table}

\begin{table}
      \centering
        \begin{tabular}{lccc}
            \toprule
            Blocks & parameters &  CD $\downarrow$ & P2S $\downarrow$\\
            \midrule
                {6} & {8.26M} & {22.81} & {0.0073}\\
                {9} & {12.17M}&{8.9} & {0.0053}\\
                {18} & {22M}& \textbf{7.71} & \textbf{0.0051}\\
            \bottomrule
        \end{tabular}
        \caption{\textbf{Ablation Study:} Effect of ResNet blocks.}
        \label{tab:resnet-blocks}
\end{table}
\begin{table}
      \centering
        \begin{tabular}{lccc}
            \toprule
            Network &  CD $\downarrow$ & P2S $\downarrow$\\
            \midrule
                {Addition}  & {8.24} & {0.0052}\\
                {Average}  & {8.82} & {0.0058}\\
                {Concat} & \textbf{7.57} & \textbf{0.0049}\\
                {Ours*} & {7.71} & {0.0051}\\
            \bottomrule
        \end{tabular}
        \caption{\textbf{Ablation Study:} Comparison of various Fusion Strategies. Ours is only end-to-end trainable mechanism as opposed to Addition, Average and Concat fusion.}
        \label{tab:fusion}
\end{table}
\begin{figure}[tbh]
        \includegraphics[width=\linewidth]{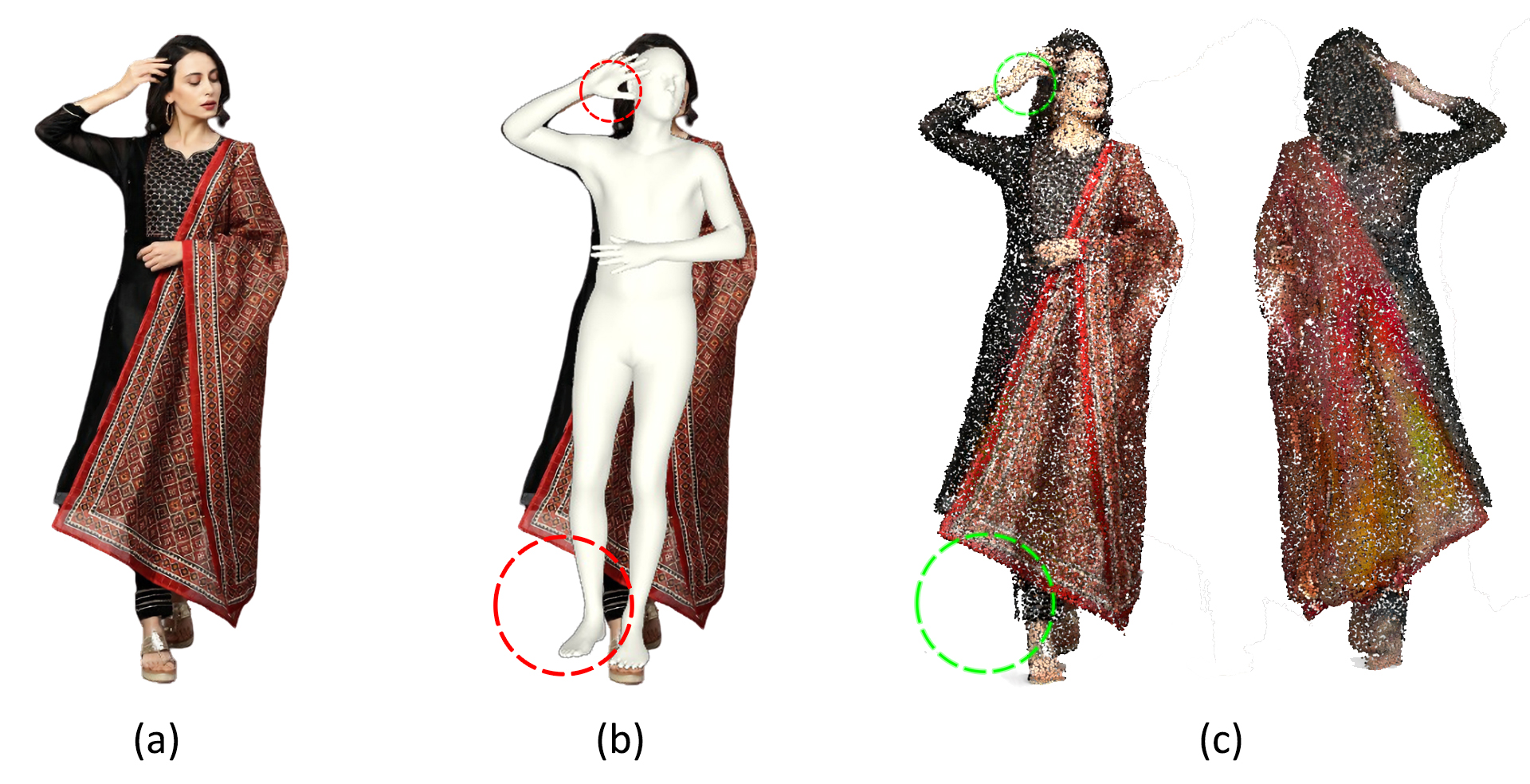}
        \caption{\textbf{Handling noisy shape prior:} (a) Input image, (b) SMPL prior misaligned with the input image, (c) Point cloud output from SHARP.}
        \label{fig:misalignment-handling}
    \end{figure}
\section{Discussion}
In this section, We perform ablative studies on various components of the network. We run all experiments on our proposed 3DHumans dataset. We also discuss in detail about the post-processing steps, along with the limitations and failure cases of our method.
\subsection{Ablation Study: Architectural Choices}
\textbf{Impact of loss functions:}   In \autoref{table:loss}, we demonstrate the impact of various loss functions on the output point cloud. 
First, we evaluate SHARP without smoothness loss ($L_{sm}$) and observe that it leads to an increase in Chamfer distance and P2S error, which is caused by the noise in prediction of fused peel maps.\\
Secondly, to evaluate the importance of the peel map fusion, we train our network without fusion. In this setting, we used only two decoder branches, one for predicting RGB peel maps and the other for predicting depth peel maps. This lead to smooth, predictions, which misses out body-specific geometrical details, further increasing CD and P2S values.\\
\\
\textbf{Impact of various backbone networks:}
We evaluate the performance of SHARP on various backbone network architectures. In particular, we used U-Net \citep{ronneberger2015u} and stacked hourglass network \citep{newell2016stacked} as backbone networks along with residual networks. All the backbone networks are trained with same loss functions as described in \autoref{subsec:losses}. We report the performance of these networks in \autoref{tab:architectures}. Residual network outperforms both Unet and hourglass networks in this multi-branch prediction task.  We also observed that hourglass network was not able to predict four layer RGB peel maps. \\
\\
\textbf{Impact of number of ResNet blocks:}
We also evaluate the performance of SHARP by varying the number of ResNet blocks as shown in \autoref{tab:resnet-blocks}.
We train our network on 3DHumans with 6, 9 and 18 blocks. Using only $6$ ResNet blocks, which is almost one-third of the original network, SHARP is able to achieve similar performance as PIFu (please refer \autoref{tab:comparison_ours}).
Using $9$ ResNet blocks, we are able to achieve closer numbers to majority of existing SOTA methods.
We observed that the further increase in the number of ResNet blocks did not yield any significant improvement.\\
\\
\textbf{Fusion Strategies}
We analyse the performance of SHARP with various fusion strategies of peel maps. In this experiment, we perform feature level fusion instead of auxiliary and residual depth peel map fusion. We train this fusion network in a coarse-to-fine strategy where initially we replace the auxiliary peel map branch with predicting complete depth peel maps $\widehat{D}_{peel}$. We train this network with losses $L_{rd}$, $L_{sm}$ and $L_1$ loss on predicted and ground truth peel maps. 
We then, take this network as initialization to train fusion module where we take intermediate features of $\widehat{D}_{peel}$ and $\widehat{D}_{rd}$ branches respectively. Refer supplementary (Figure 4) for the architecture diagram. We fuse them using three strategies (a) addition, (b) average and (c) concatenation. These fused features are then passed to upsampling and convolutional layers to predict final fused depth peel maps. Here, we freeze the weights of the network except the layers after the feature fusion. We call it as \textit{Late Fusion} as it requires pre-trained network. 

We report the performance in \autoref{tab:fusion} and learn that late fusion with concatenation results in better performance. However, we note the training for late fusion is not end-to-end as described and we adopt end-to-end trainable network with fusion proposed in \autoref{eq:fusion_eq} as our final choice.\\
\\

\subsection{{Handling Noisy Shape Prior}}
The shape prior based reconstruction methods are susceptible to noisy initialization from incorrect prior. 
Generally, this leads to incorrect pose conditioning, which further deteriorates the final reconstruction.  
Our method can partially handle such noisy prior 
as we use refined per-layer SMPL mask $\Gamma_i$ (introduced in \autoref{subsubsec:peeled_shape_prior.}, \autoref{eq:SMPL-mask}) to mask out the regions of the SMPL prior peel maps which fall outside the clothed human silhouette in input image. Thus, residual deformation $\widehat{\mathcal{D}}_{rd}$ predicted for the misaligned regions of the SMPL prior is not considered during fusion, and we are able to avoid the errors in reconstruction due to such misalignments. 
\autoref{fig:misalignment-handling} shows a case of noisy SMPL prior for an in-the-wild image and the final reconstruction output of SHARP, where it is able to recover from incorrect prior in the leg and hand region.
\begin{figure*}
    \centering
        \includegraphics[width=0.9\linewidth]{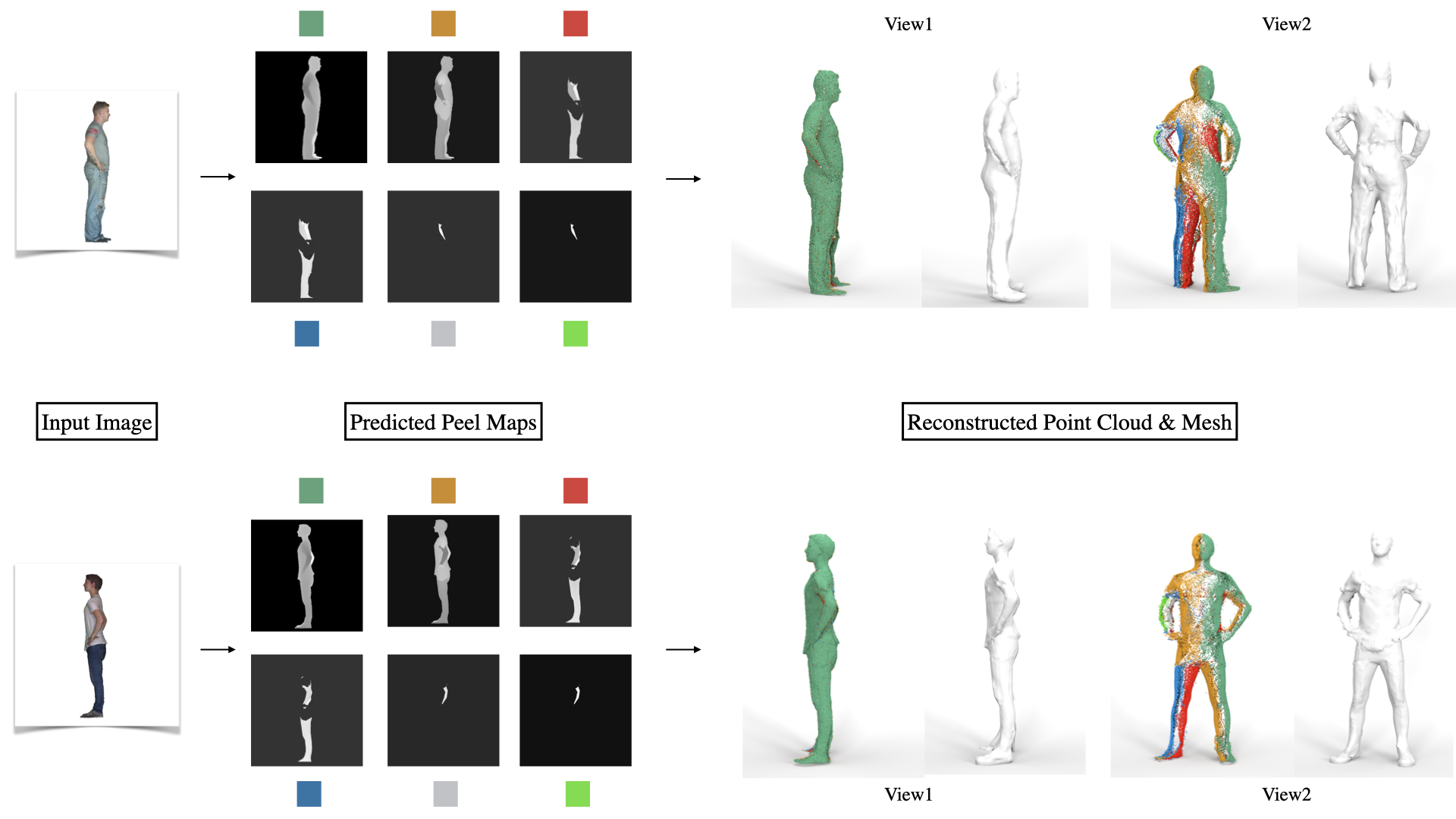}
        \caption{Performance of our method on six peel layer representation. We show the predicted final fused depth peel maps (with corresponding color coding) along with backprojected point cloud (points from a layer is color coded with the same color as indicated in depth peel maps) and reconstructed mesh respectively.}
        \label{fig:six_layer}
    \end{figure*}
\subsection{Peeled Representation Layers}
In this work, we used four layers of peeled representation for human body recovery. However, our formulation is generalizable to arbitrary number of layers. In \autoref{fig:six_layer}, we show the performance of SHARP on real images (not included in training data distribution) where six layers are needed to capture the geometry. Note that SHARP is able to recover geometry under the case of severe self-occlusion and with skewed viewpoints. To train this network, we initially train network with four layers  and then using this model to initialize weights for model with six layers. 
\subsection{Post-processing}
The output of our network is prone to slight noise in the predicted peel maps, resulting in sparse outliers in the back-projected point cloud, as shown in \autoref{fig:postprocess} (a). These outliers are removed by density-based filtering, where we fit spheres with 16 neighbours on each point. The points, which are inside the spheres having a radius greater than the threshold (0.01 in our case), are removed to obtain a clean point cloud, as shown in \autoref{fig:postprocess} (b). Finally, the filtered point cloud might have some small holes which are subsequently filled by meshification using Poisson Surface Reconstruction (PSR) \autoref{fig:postprocess} (c).

\subsection{Limitations}
\begin{figure}
        \includegraphics[width=\linewidth]{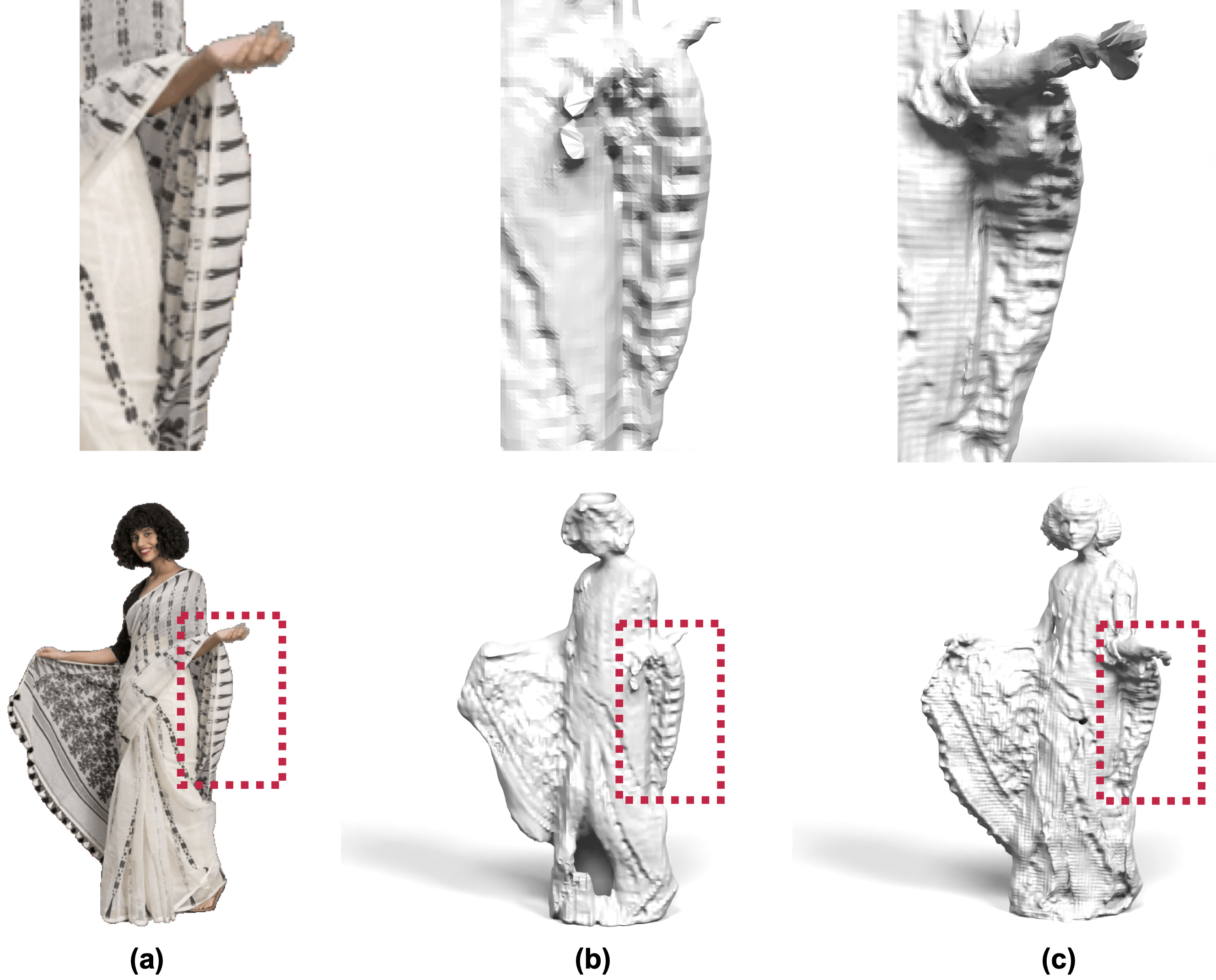}
        \caption{\textbf{Texture-Geometry Ambiguity:} High-frequency textural details can be interpreted as geometrical details by monocular deep reconstruction techniques.
        (a) Input image, (b) PaMIR and (c) SHARP.}
        \label{fig:limit}
    \end{figure}

\begin{figure}
        \includegraphics[width=\linewidth]{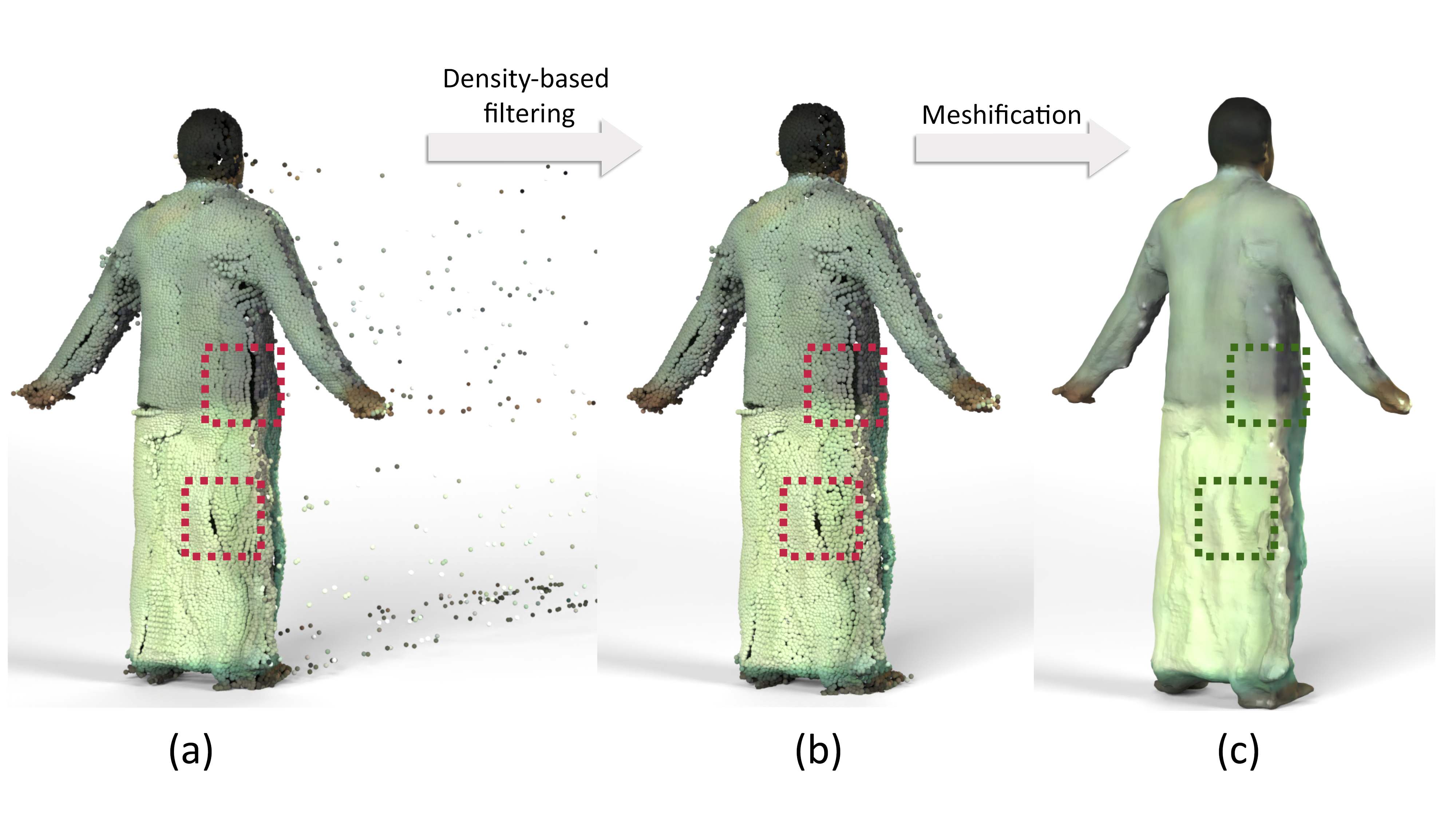}
        \caption{Effect of post-processing.}
        \label{fig:postprocess}
    \end{figure}

\begin{figure}[h!]
        \includegraphics[width=\linewidth]{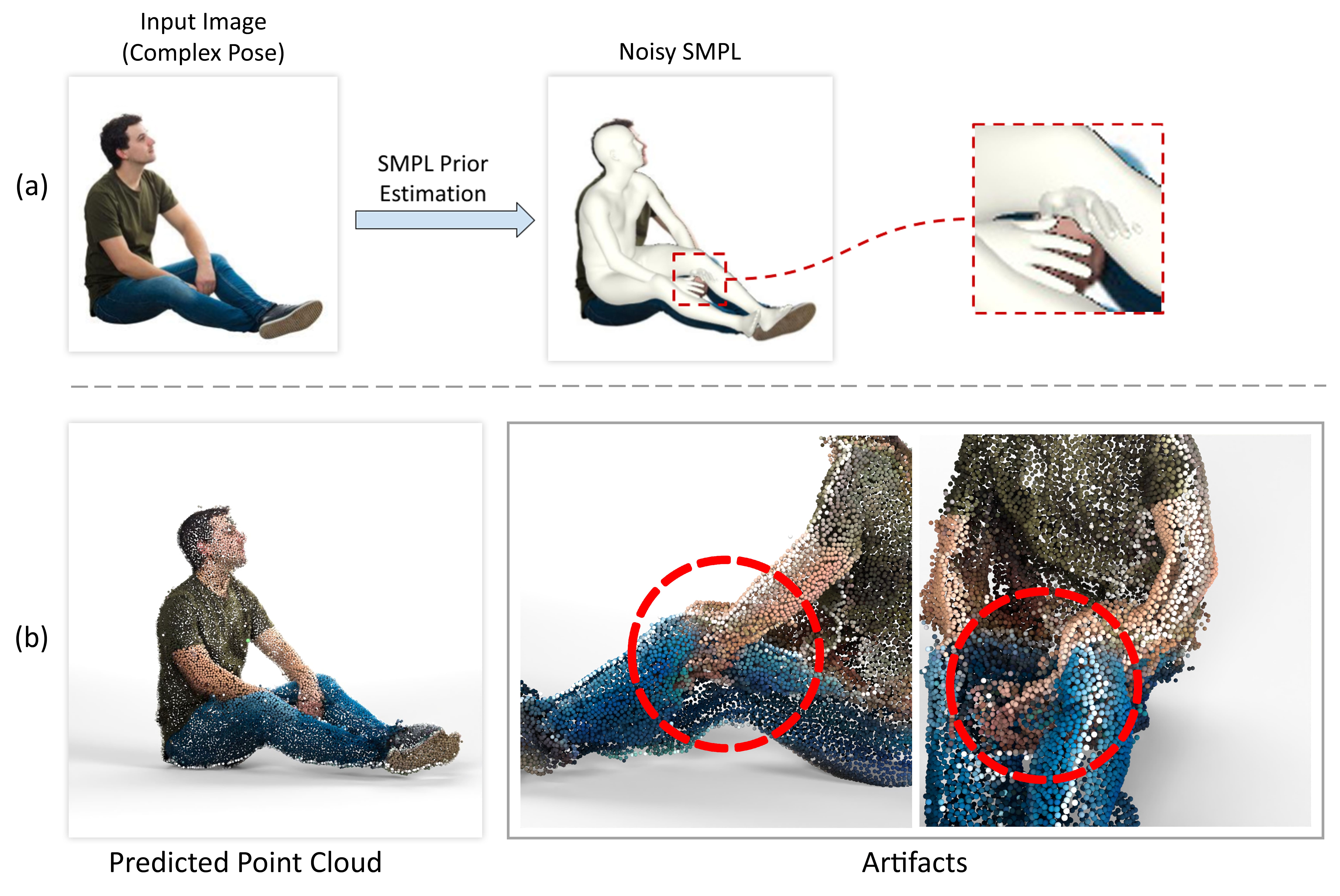}
        \caption{\textbf{Failure Case} : \textbf{(a)} Noisy SMPL estimation (hands are intersecting with the legs) due to highly complex pose. \textbf{(b)} Artifacts in the predicted point cloud.}
        \label{fig:failure_case}
    \end{figure}

\textbf{Ambiguity due to textural edges:}
3D reconstruction from a monocular RGB image, being an ill-posed problem, is susceptible to interpreting the textural edges as geometrical details. In \autoref{fig:limit}, we show reconstructions from our method and PaMIR, where both the methods incorrectly interpret textural details of a flat clothing surface as geometrical details and hallucinate geometrical structures, which are non-existent.\\
\\
\\
\textbf{Failure cases:} One of the key challenges faced by majority of existing prior-based methods is self-intersection of body parts in the prior, mainly due to challenging poses. In  \autoref{fig:failure_case}, a failure case of our approach is shown where the network reconstructs the occluded regions accurately, but fails to recover from interpenetrating body parts, present in the input SMPL prior (hands penetrating the legs).

\section{Conclusion}
Reconstructing 3D clothed human body from a monocular RGB image is an extremely ill-posed problem due to skewed viewpoints, depth ambiguities, complex poses and arbitrary clothing styles. Although many solutions exist which can recover clothed human body in relatively tighter clothing, they fail to generalize when it comes to in-the-wild loose clothing scenarios. 
To this end, we have contributed a novel end-to-end trainable deep learning framework, SHARP, which uses a sparse and efficient fusion of parametric body prior with non-parametric PeeledHuman representation, and is able to reconstruct human body in arbitrarily loose clothing.

In more general perspective, we built on our sparse non-parametric 2D shape representation and proposed an efficient strategy to fuse it with parametric shape prior. We train a compact, encode-decoder based network using a set of L1 losses on 2D maps, while reconstructing complex 3D geometry. The proposed formulation is sparse in terms of representation, resulting in low inference time of the network.

We evaluated our framework on various publicly available datasets and reported superior qualitative and quantitative performance as compared to state-of-the-art methods.
Since, data is a key bottleneck in the field of deep learning based 3D human body reconstruction, we contributed 3DHumans dataset and intend to release it in the public domain to further accelerate the research. Our dataset contains 3D human body scans of high-frequency textural and geometrical details, with a wide variety of the body shapes in various clothing styles. 

Although per-frame reconstruction of SHARP yields reasonable intra-frame consistency without any explicit temporal conditioning (as shown in the supplementary video), it will be interesting to explore extension of our method to learn over video sequences where it is difficult to get high quality ground-truth data. Another interesting direction is to incorporate learning from multi-view images for better reconstruction results. Additionally, performance of our method can be further improved by addressing the texture-geometry ambiguity and recovering from challenging scenarios such as self-intersecting body parts.


\newpage
\bibstyle{basic} 


\end{document}